\pdfoutput=1

\documentclass[11pt]{article}

\usepackage[final]{acl}

\usepackage{times}
\usepackage{latexsym}
\usepackage{graphicx}
\usepackage{subcaption}
\usepackage{amssymb}

\usepackage[T1]{fontenc}

\usepackage[utf8]{inputenc}

\usepackage{microtype}

\usepackage{inconsolata}

\usepackage{graphicx}
\usepackage{tabularx}
\usepackage{booktabs} 
\usepackage{longtable} 
\title{Evaluating Vision-Language Models on Bistable Images}

\author{
 \textbf{Artemis Panagopoulou$^\ast$},
 \textbf{Coby Melkin$^\ast$},
 \textbf{Chris Callison-Burch}\\
 University of Pennsylvania\\
 \small{
   \textbf{Correspondence:} \href{mailto:artemisp@seas.upenn.edu}{artemisp@seas.upenn.edu}
 }
}

\begin{document}

\maketitle
\begin{abstract}
Bistable images, also known as ambiguous or reversible images, present visual stimuli that can be seen in two distinct interpretations, though not simultaneously by the observer. In this study, we conduct the most extensive examination of vision-language models using bistable images to date. We manually gathered a dataset of 29 bistable images, along with their associated labels, and subjected them to 116 different manipulations in brightness, tint, and rotation. We evaluated twelve different models in both classification and generative tasks across six model architectures. Our findings reveal that, with the exception of models from the Idefics family and LLaVA1.5-13b, there is a pronounced preference for one interpretation over another among the models, and minimal variance under image manipulations, with few exceptions on image rotations. Additionally, we compared the models’ preferences with humans, noting that the models do not exhibit the same continuity biases as humans and often diverge from human initial interpretations. We also investigated the influence of variations in prompts and the use of synonymous labels, discovering that these factors significantly affect model interpretations more than image manipulations showing a higher influence of the language priors on bistable image interpretations compared to image-text training data. All code and data is open sourced~\footnote{\href{https://github.com/artemisp/Bistable-Illusions-MLLMs.git}{\fontsize{8}{8}\selectfont https://github.com/artemisp/Bistable-Illusions-MLLMs.git}}.

\renewcommand{\thefootnote}{*}  
\renewcommand{\thefootnote}{\arabic{footnote}}

\end{abstract}

\section{Introduction}
\begin{figure}[ht]
  \includegraphics[width=\columnwidth]{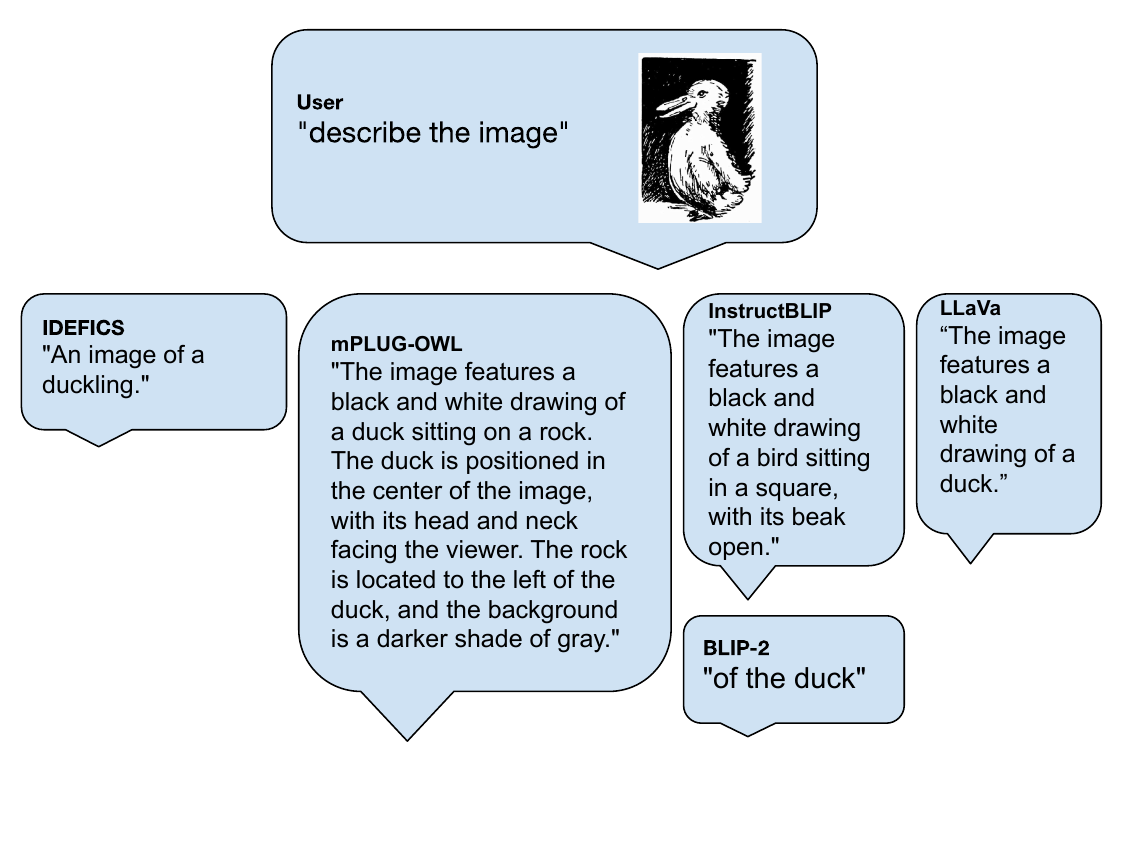}
  \vspace{-1cm}
  \caption{Depiction of generative models' descriptions of a Duck-Rabbit image. Responses are drawn directly from model outputs.}
  \label{fig:generative_example}
\end{figure}

Bistable images, also known as ambiguous or reversible images, offer unique visual stimuli that present two distinct interpretations, though a viewer cannot simultaneously perceive both\citep{khalil2021does}. An example of this is depicted in Figure \ref{fig:duck_rabbit_gen}, which can be seen as either a rabbit or a duck. The rapid advancements in vision-language models~\citep{ye2023mplug, radford2021learning, dai2023instructblip, liu2023visual, li2023blip} have sparked interest in testing these models against various types of visual challenges, including optical illusions. While considerable research has been done on how these models interpret geometric and color-varying optical illusions~\citep{guan2023hallusionbench, gomez2019convolutional, zhang2023grounding, afifi2019else, benjamin2019shared, sun2021imagenet}, exploration into their performance with bistable images remains sparse.

Motivated by this gap, this work aims to conduct a comprehensive investigation into how vision-language models process and interpret bistable images. We assemble the largest dataset of bistable images to date, apply a range of visual transformations, and examine the models' interpretations and their alignment with human perception. 

In particular, we collect 29 bistable images from diverse online sources and cognitive science literature. Each image is subjected to 116 transformations affecting brightness and tint, resulting in a total of 3,364 processed images. We assessed the behaviors of twelve vision-language models across six distinct model families in both classification and generative settings. Our analysis shows that, apart from a few exceptions, these models generally demonstrate a preference for one interpretation of bistable images over the other. Notably, models from the Idefics family~\citep{laurenccon2024obelics} and LLaVA1.5-13b~\cite{liu2023visual,liu2023improved} exhibit more balanced preferences. Additionally, while most model responses show little variation to image manipulations, exceptions include CLIP~\citep{radford2021learning} and BLIP2 OPT6.7~\citep{li2023blip}, which are sensitive to such changes.

To further understand the influence of training data, we considered multiple models from the same families, trained on identical datasets but using different base language models (LLMs). This approach revealed that even when trained on the same visual data, the models do not consistently align in their preferences, suggesting that LLM priors play a major role in ambiguous image interpretation. 

Additionally, we explored how variations in prompts and the use of synonymous labels affect model interpretations. These textual modifications significantly influenced the models’ interpretations, reinforcing the importance of LLM priors on the processing of bistable images by vision-language models. This finding contrasts with previous research on convolutional neural networks (CNNs) focused on geometric optical illusions~\citep{gomez2019convolutional, gomez2020color, afifi2019else, benjamin2019shared, sun2021imagenet}, which typically show biases consistent with human perception. The CNNs studied did not utilize language model priors, highlighting a fundamental difference in how traditional vision models and VLMs handle visual ambiguity.

\noindent Our contributions are as follows:
\begin{itemize}
\vspace{-.3cm}
\item We have curated the largest collection of bistable images from various online sources and cognitive studies, consisting of 29 unique images. These images have been modified through 116 transformations, creating a comprehensive set of 2.3k images for analysis.
\vspace{-.3cm}
\item We analyze the behavior of twelve different vision-language models across six architectural types in both classification and generative tasks, providing a detailed account of their performance on bistable images.
\vspace{-.3cm}
\item We examine the influence of prompt variations and synonymous labeling on model interpretations, finding that these textual modifications significantly impact how models perceive bistable images.
\vspace{-.3cm}
\item Through direct comparison with human subjects and reference to established cognitive science studies, we assess the degree to which model preferences align with humans. Interestingly, we find that unlike previous work on CNNs~\citep{gomez2019convolutional, gomez2020color, afifi2019else, benjamin2019shared, sun2021imagenet}, VLMs do not exhibit human biases in bistable images interpretations.
\end{itemize}

\section{Background}
\subsection{Bistable Images}
Bistable images, a unique class of cognitive illusions, present two or more plausible perceptual states, yet viewers cannot observe multiple percepts simultaneously \citep{khalil2021does}. Instead, observers typically "switch" between the percepts in a seemingly random manner \citep{KORNMEIER2005955}. This phenomenon prompts two primary questions in Cognitive Science regarding bistable images:
\begin{enumerate}
\vspace{-.3cm}
\item What causes an individual to initially perceive a particular percept?
\vspace{-.3cm}
\item What triggers the seemingly random switching between percepts?
\vspace{-.3cm}
\end{enumerate}

The exploration of these questions incorporates both bottom-up and top-down considerations \citep{wang2013brain}. Bottom-up explanations focus on how the brain processes visual stimuli, starting from the simplest sensory inputs and moving to more complex interpretations. This process involves the detection of subtle visual cues and the neural computation within the visual cortex that ultimately determines the perceived image. Conversely, top-down explanations emphasize the role of cognitive processes, such as expectations, which heavily influence initial perceptions. For instance, a person's previous experiences, like frequently viewing cubes from above, shape their initial interpretation of a Necker Cube \citep{kuc2023studying}.

Regarding the switching phenomenon, the dominant bottom-up explanation involves neural mechanisms like spike frequency adaptation or synaptic depression, where the neural connections producing one percept become fatigued, allowing the alternative percept to emerge \citep{laing2002spiking}. Other bottom-up theories propose that this switching is influenced by the brain’s inherent noise or randomness \citep{moreno2007noise} or by unconscious, subtle cues within the images \citep{ward2015stochastic}. On the other hand, top-down explanations suggest that higher cognitive functions, such as motivation and attention, can also induce switching. Studies have shown that individuals can exert some control over their perceptual focus, which influences the switching between different states \citep{hugrass2012willpower, slotnick2005common}.

\subsection{Vision-Language Models (VLMs)}
Vision-Language Models (VLMs) integrate visual information as input and generate text as output. VLMs are categorized into contrastive and generative types. Contrastive VLMs, such as the prototypical model CLIP \cite{radford2021learning}, are trained to match visual representations with corresponding textual descriptions by distinguishing between different data points. These models create a latent embedding space where similar text and images are drawn closer together, while dissimilar ones are pushed apart. Generative VLMs extend this by incorporating a vision-to-language connection module that projects visual information into the LLM space. This module can either prepend to the input layer of the LLM or condition deeper layers through cross-attention. The integration allows for flexible and dynamic text generation based on visual inputs. For our experiments, we employed models from various families, including CLIP, Idefics \citep{laurenccon2024obelics}, LLaVA1.5 \citep{liu2023visual, liu2023improved}, mPLUG-Owl \citep{ye2023mplug}, InstructBLIP \citep{dai2023instructblip}, and BLIP-2 \citep{li2023blip}. Detailed information on the model architectures and the datasets used for training these models is presented in the appendix, in Tables \ref{tab:models} and \ref{tab:model_data}.

\subsection{VLMs and Cognitive Illusions}
While prior studies have investigated how Convolutional Neural Networks (CNNs) process optical illusions, showing that they often mimic human perceptual errors \citep{gomez2020color, gomez2019convolutional, afifi2019else, benjamin2019shared, sun2021imagenet}, the interaction of VLMs with cognitive illusions, especially bistable images, remains underexplored. More closely related to this work, \citet{zhang2023grounding} evaluated VLMs on optical illusions by soliciting binary Yes/No responses and found that larger VLMs tend to be more susceptible to such illusions. However, their study was limited to 16 root images with 100 manually edited variations, focusing primarily on color, shape, and geometric illusions and did not include bistable images. Furthermore, they experimented with only three families of models, whereas our study encompasses six. Limited resources restricted our ability to test some of the larger models that \citet{zhang2023grounding} included. Hallusion-bench \citep{guan2023hallusionbench} integrates a subset of these optical illusion images, predominantly sourced from \citet{zhang2023grounding}, but lacks bistable examples.

\section{Methodology}

\subsection{Data Collection}
Our dataset comprises 29 bistable images categorized into seven distinct types, sourced from both online platforms, such as Wikipedia, and academic studies, notably from the \citet{takashima2012face} research on face perception illusions. Among these, twelve images are organized into four classic categories of bistable illusions: the Rubin Vase, Necker Cube, Duck-Rabbit, and Young-Old Woman. Each category includes several iconic versions of the respective illusion type.

To explore the influence of visual modifications on perception, we created 116 variations for each image through a series of controlled manipulations. These manipulations include adjustments to image brightness—both increases and decreases—and the application of color tints. The specific colors used for the tints, along with their RGB values, are as follows: \textcolor{red}{red}, \textcolor{green}{green}, \textcolor{blue}{blue}, \textcolor{yellow}{yellow}, \textcolor{magenta}{magenta}, and \textcolor{cyan}{cyan}. The intensity of each tint was varied by 0.1 from 0 (no change) to 1.0 (maximum change), and the brightness was adjusted within a range from -1 (darker) to 1 (brighter). We also applied image rotations from 0 to 360 degrees every 10 degrees.

\subsection{Experimental Setup}
We utilized six VLM families, encompassing a total of twelve different models, to evaluate bistable image description. We employed all six VLMs for classification tasks and five for generation tasks (excluding CLIP). The models used and their corresponding implementations on Huggingface Transformers are listed in the footnotes: CLIP~\citep{radford2021learning}\footnote{\fontsize{7}{7}\selectfont\href{https://huggingface.co/openai/clip-vit-base-patch32}{\texttt{openai/clip-vit-base-patch32}}, \href{https://huggingface.co/openai/clip-vit-base-patch16}{\texttt{openai/clip-vit-base-patch16}}, \href{https://huggingface.co/laion/CLIP-ViT-B-32-laion2B-s34B-b79K}{\texttt{laion/CLIP-ViT-B-32-laion2B-s34B-b79K}}}, Idefics 9b~\citep{laurenccon2024obelics}\footnote{\fontsize{7}{7}\selectfont\href{https://huggingface.co/HuggingFaceM4/idefics-9b}{\texttt{HuggingFaceM4/idefics-9b}}, \href{https://huggingface.co/HuggingFaceM4/idefics-9b-instruct}{\texttt{HuggingFaceM4/idefics-9b-instruct}}}, LLaVA1.5~\citep{liu2023visual,liu2023improved}\footnote{\fontsize{7}{7}\selectfont\href{https://huggingface.co/llava-hf/llava-1.5-7b-hf}{\texttt{llava-hf/llava-1.5-7b-hf}}, \href{https://huggingface.co/llava-hf/llava-1.5-13b-hf}{\texttt{llava-hf/llava-1.5-13b-hf}}}, mPLUG-Owl~\citep{ye2023mplug}\footnote{\fontsize{7}{7}\selectfont\href{https://huggingface.co/MAGAer13/mplug-owl-llama-7b}{\texttt{MAGAer13/mplug-owl-llama-7b}}}, InstructBLIP~\citep{dai2023instructblip}\footnote{\fontsize{7}{7}\selectfont\href{https://huggingface.co/Salesforce/instructblip-flan-t5-xl}{\texttt{Salesforce/instructblip-flan-t5-xl}}}, and BLIP-2~\citep{li2023blip}\footnote{\fontsize{7}{7}\selectfont\href{https://huggingface.co/Salesforce/blip2-opt-2.7b}{\texttt{Salesforce/blip2-opt-2.7b}}, \href{https://huggingface.co/Salesforce/blip2-opt-6.7b}{\texttt{Salesforce/blip2-opt-6.7b}}, \href{https://huggingface.co/Salesforce/blip2-flan-t5-xl}{\texttt{Salesforce/blip2-flan-t5-xl}}}. Each model was queried with the default generation parameters and the prompt suggested by their respective model page on Huggingface. All experiments were conducted on a single A100 40GB GPU.

Although all VLMs used, except for CLIP, are generative models, we adapted their outputs to simulate classification. Specifically, we utilized a loss ranking technique~\citep{wei2021finetuned,li2021align,li2023blip,dai2023instructblip} for classification, employing the score to determine the negative log likelihood of each candidate label. In the classification setup, we prompted each VLM with each image along with a pair of strings corresponding to its potential interpretations\footnote{Image interpretations are found in Appendix \ref{app:collection}}.

In the generative setup, we prompted the models with the format suggested in the HuggingFace documentation for captioning. In addition to model-specific setups, all models were presented with each image and asked to \texttt{\small ``describe the image."} 

\section{Results}

\subsection{VLMs on Original Images}
\begin{figure}[t]
\begin{subfigure}[b]{\columnwidth}
\centering
      \includegraphics[width=\columnwidth
]{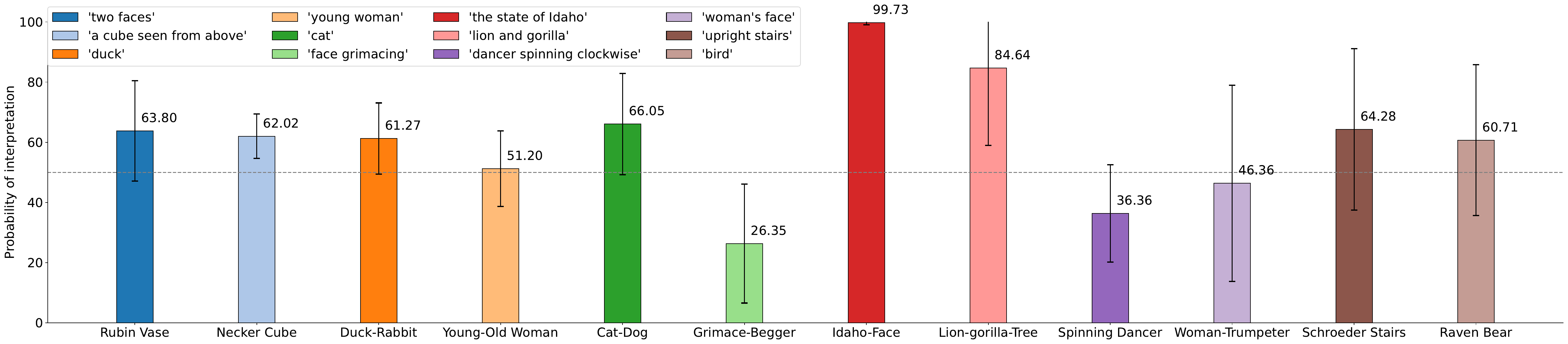}
  \caption{Original Labels}
    \label{fig:original_averages}

\end{subfigure}
\begin{subfigure}[b]{\columnwidth}
\centering
 \includegraphics[width=\columnwidth]{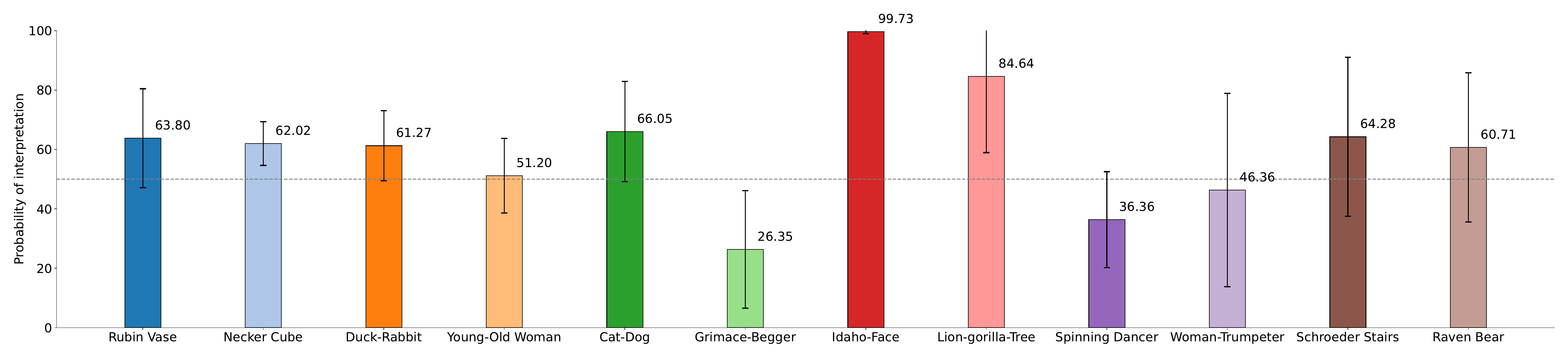}
  \caption{Synonym Labels}
  \label{fig:syn_averages}
\end{subfigure}

\begin{subfigure}[b]{\columnwidth}
\centering
 \includegraphics[width=\columnwidth
]{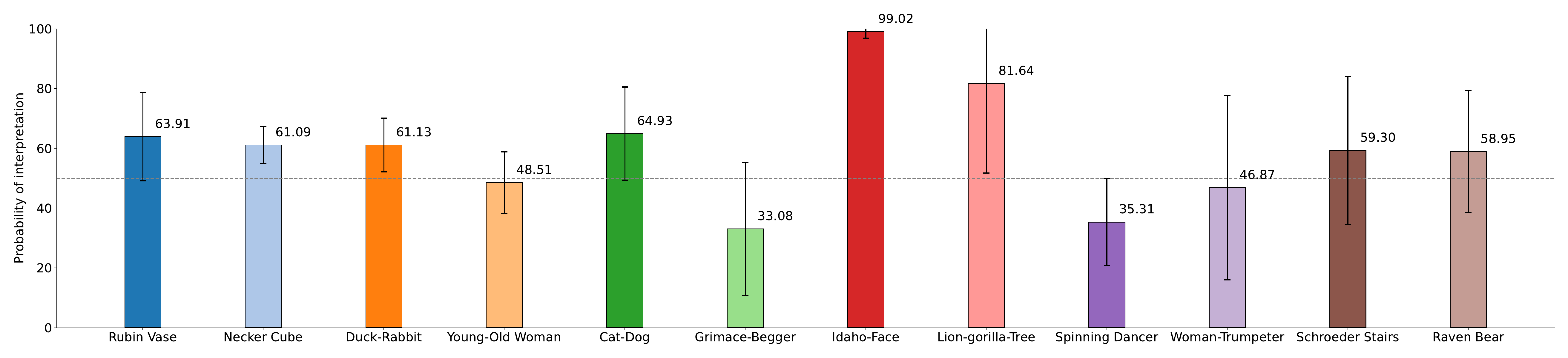}
  \caption{Prompt Variation with Original Labels}
    \label{fig:prompt_averages}
\end{subfigure}

\begin{subfigure}[b]{\columnwidth}
\centering
 \includegraphics[width=1\columnwidth
]{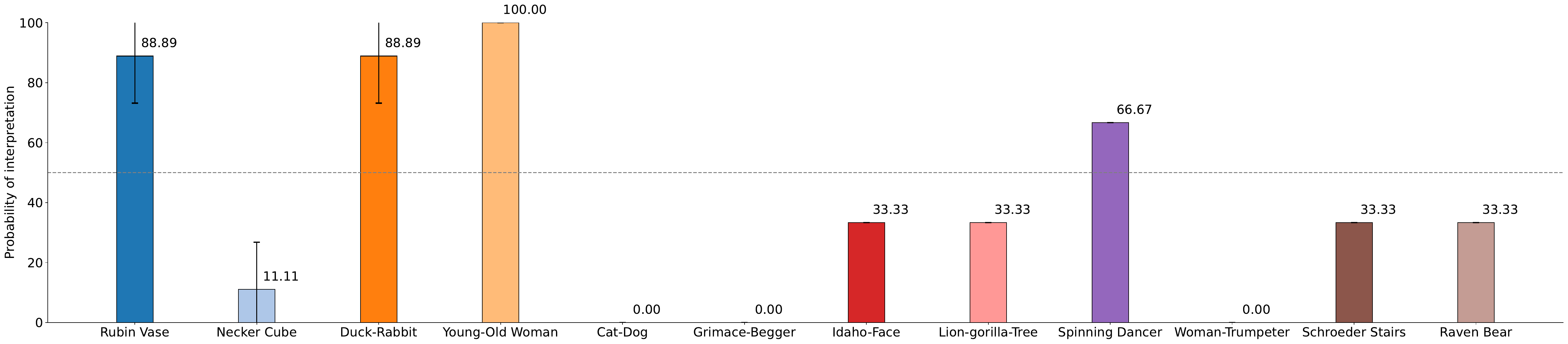}
  \caption{Human Initial Interpretations}
    \label{fig:human_averages}
\end{subfigure}
\caption{Between-model averages of probability of the favored interpretation for each image category.}
  \label{fig:between-model_averages}
\end{figure}

The models displayed clear preferences between interpretations for the original bistable images. Very rarely were models indifferent between interpretations. The averages between models for our four image categories are shown in Figure \ref{fig:between-model_averages}. We see a strong preference for the `two faces' interpretation in the Rubin Vase group moderate preferences for `a cube seen from above' and `duck' interpretations in Necker Cube and Duck-Rabbit groups. Less classical illusions such at the `Grimace-Begger, `Idaho-face', and `Lion-Gorilla-Tree` also show strong inclinations towards one interpretation. The images with the highest variation across models where the `Woman-Trumpeter', `Schroeder Stairs', and `Raven-Bear' with CLIP variants showing almost consistently opposite preferences to the LLM based generative models.

While the six models generally showed alignment in their interpretation preferences, there was significant variance observed. Figure \ref{fig:models_averages} in the Appendix displays the probability distributions for each image category across individual models, revealing some noteworthy model-specific trends. Firstly, all CLIP variants exhbited the exact same probability distributions, with high variance across images within the same category, suggesting a heightened sensitivity to bistability. Secondly, the variants of Idefics 9b and LLaVA 13b demonstrated minimal variance among images of the same category and exhibited relatively moderate preferences, indicating a lower sensitivity to bistability. Interestingly, all models showed a preference for the two animals over the tree in the `Lion-Gorilla-Tree' illusion, despite the frequent appearance of all these objects in their training sets. Additionally, there was a consistent preference for the face over the full-body abstract silhouette in the   `Grimace-Begger' illusion across all models, except those based on the Flan T5xl architecture. This highlights the significant impact of the underlying LLM on image interpretation in VLMs. Notably, although BLIP2 OPT was trained on the same image-text data as the Flan T5 variants, it exhibited almost opposite preferences in some image categories.

\subsection{VLMs on Image Manipulations}

\begin{figure*}
\centering
\begin{subfigure}[b]{\textwidth}
\centering
      \includegraphics[width=\linewidth
]{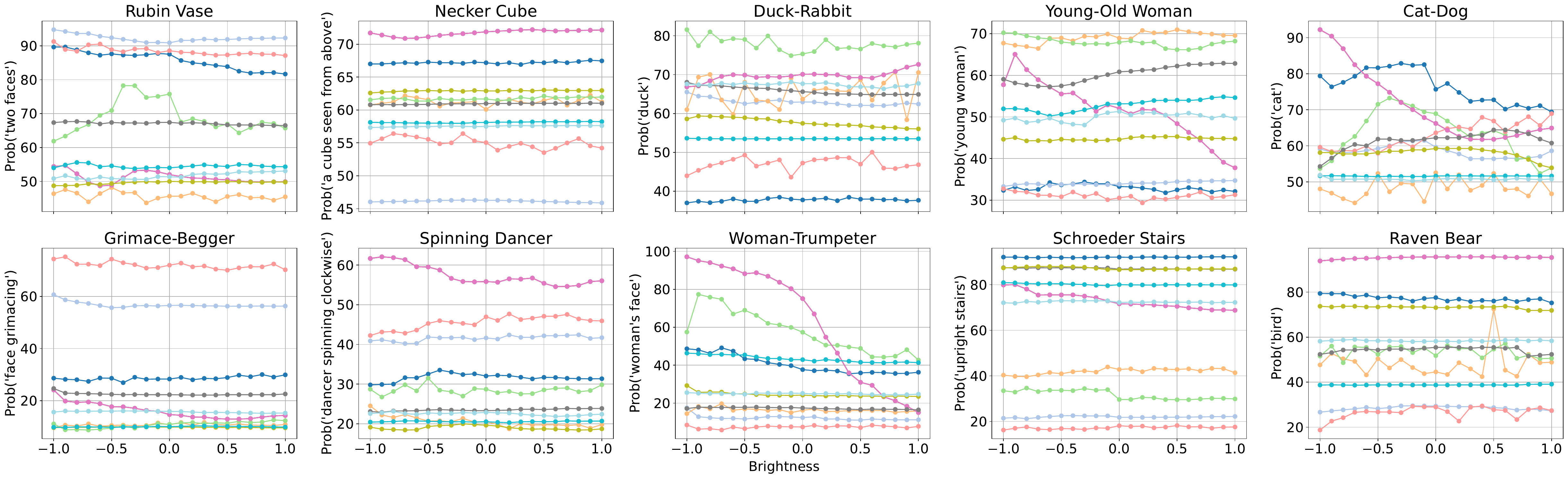}
  \caption{Brightness variation.}
\label{fig:brightness}
\end{subfigure}
\begin{subfigure}[b]{\textwidth}
\centering
\includegraphics[width=\linewidth
]{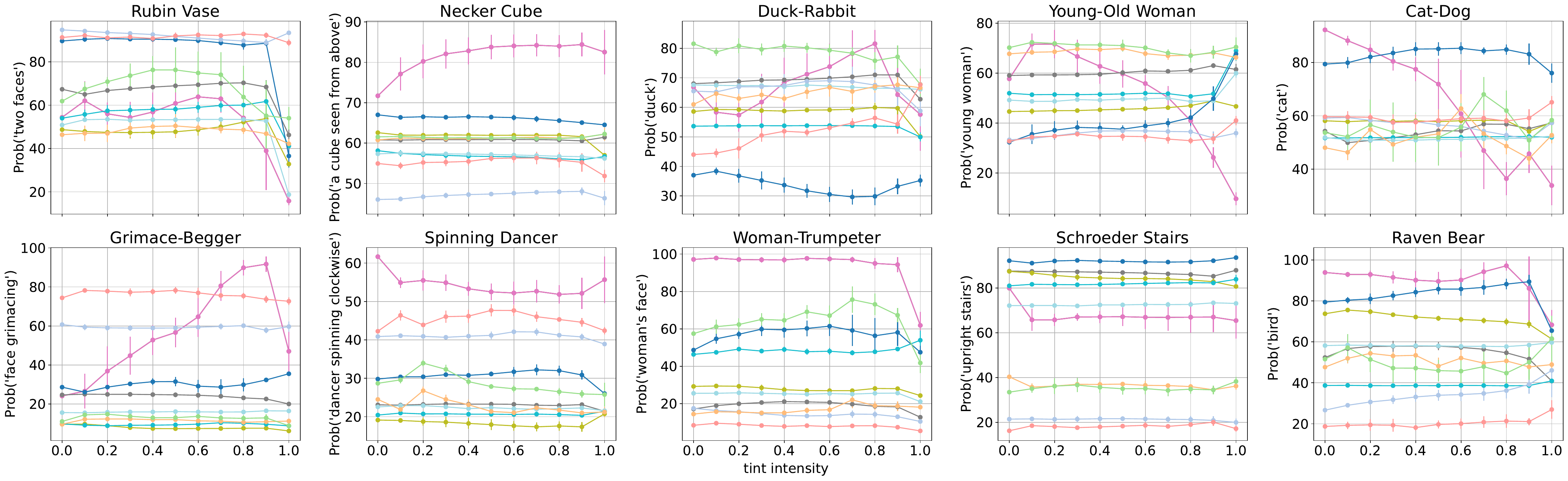}
  \caption{Tint variation. Average across six color tints.}
  \label{fig:tint}
\end{subfigure}

\begin{subfigure}[b]{\textwidth}
\centering
\includegraphics[width=\linewidth
]{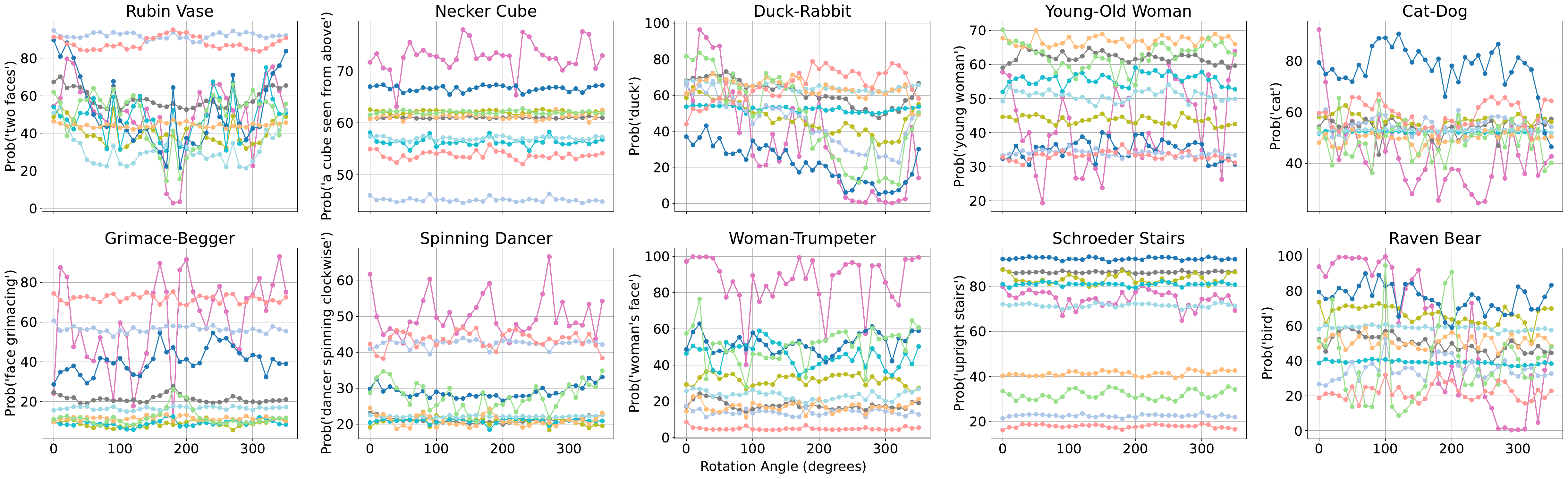}
  \caption{Rotation Variation.}
  \label{fig:rotation}
\end{subfigure}
\begin{subfigure}[b]{\textwidth}
\centering
\includegraphics[width=\linewidth
]{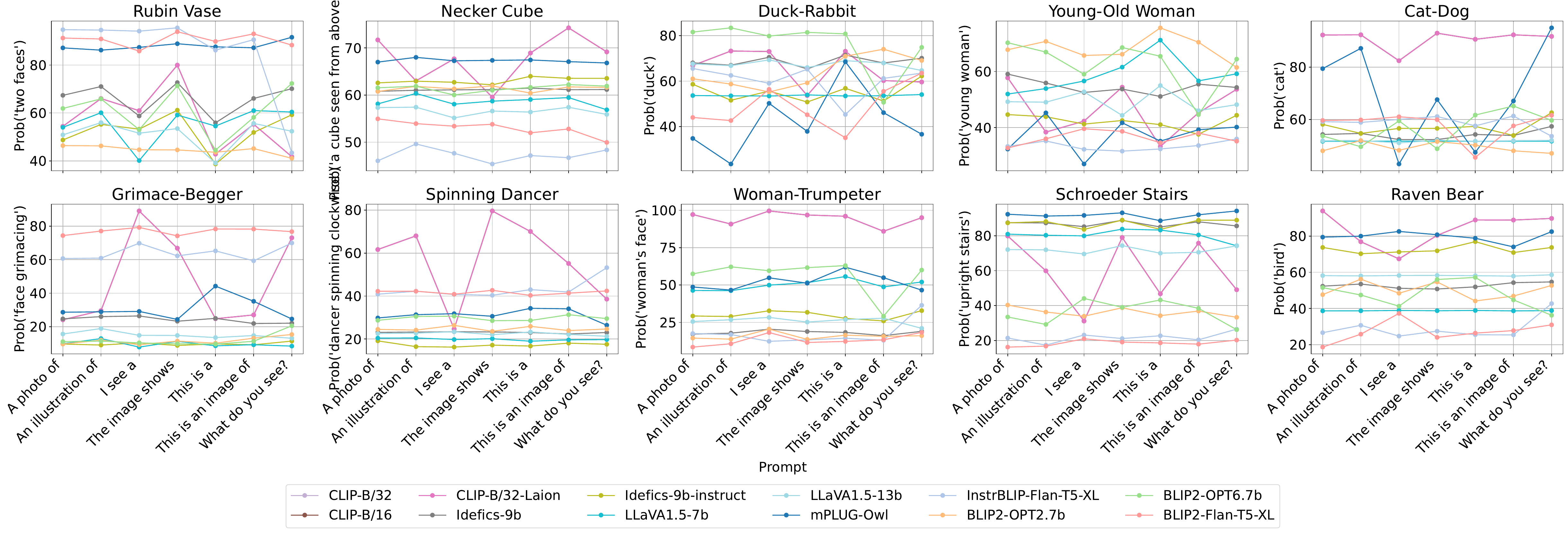}
  \caption{Prompt variation}
  \label{fig:prompt}
\end{subfigure}
\caption{Bistable image interpretation under brightness (a), tint (b), and prompt (c) manipulations.}
\label{fig:interventions}
\end{figure*}
We observed minimal effects from image manipulations on interpretation probabilities. When adjusting brightness levels, color tints, and tint intensities, the probabilities for each model remained largely unchanged. Figures \ref{fig:brightness} and \ref{fig:tint} illustrate the minimal impact of these manipulations on model interpretations. This suggests that VLMs tend to overlook minor, low-level perturbations in favor of holistic image processing. Moreover, this finding highlights a significant divergence between VLM processing and human perception of bistable images, which often relies on bottom-up cues according to certain theories \citep{ward2015stochastic}. Notably, the models did not shift interpretations based on subtle cues of brightness and color. The primary exception was the CLIP variants, which demonstrated sensitivity to variations in brightness and tint, particularly in the `Young-Old Woman,' `Cat-Dog,' `Grimace-Begger,' and `Woman-Trumpeter' illusions. We hypothesize that contrastive learning across aggregation of  patches in these models enhance their sensitivity to global changes in the image, as each layer encompasses a more substantial portion of the visual input, making any variations more influential to the model's output. This sensitivity was also observed, though to a lesser extent, in BLIP2-OPT6.7, especially regarding brightness changes in the `Rubin-Vase' and `Woman-Trumpeter' illusions. These variations were less pronounced in BLIP2-OPT2.7, particularly for the `Duck-Rabbit' illusion, and were absent in the corresponding FlanT5-xl variant, underscoring the impact of the underlying LLM's priors on generative vision-language models. Interestingly, when transformations were applied at maximum scale, resulting in a monochrome image, most models exhibited similar preferences, reinforcing the role of language priors in their processing.

Figure \ref{fig:rotation} shows the variation of interpretations across rotated versions of the images. We find that this manipulation causes significantly higher variation to the color-based manipulations. The variations   typically follow the same pattern across models for some bistable images, such as `Rubin-Vase' and `Duck-Rabbit'. Notably, contrastive based CLIP-variants once again exhibit the most variation despite being trained with `minor rotations' data augmentations. From the generative models mPLUG-Owl seems to exhibit the highest sensitivity to rotation despite also employing rotation augmentation in training. We also observe that the larger LLM variants of LLaVA1.5 and BLIP2-OPT exhibit less variation compared to their smaller counterparts, likely due to the stronger language prior. 


\begin{figure}[ht]
  \includegraphics[width=.9\columnwidth]{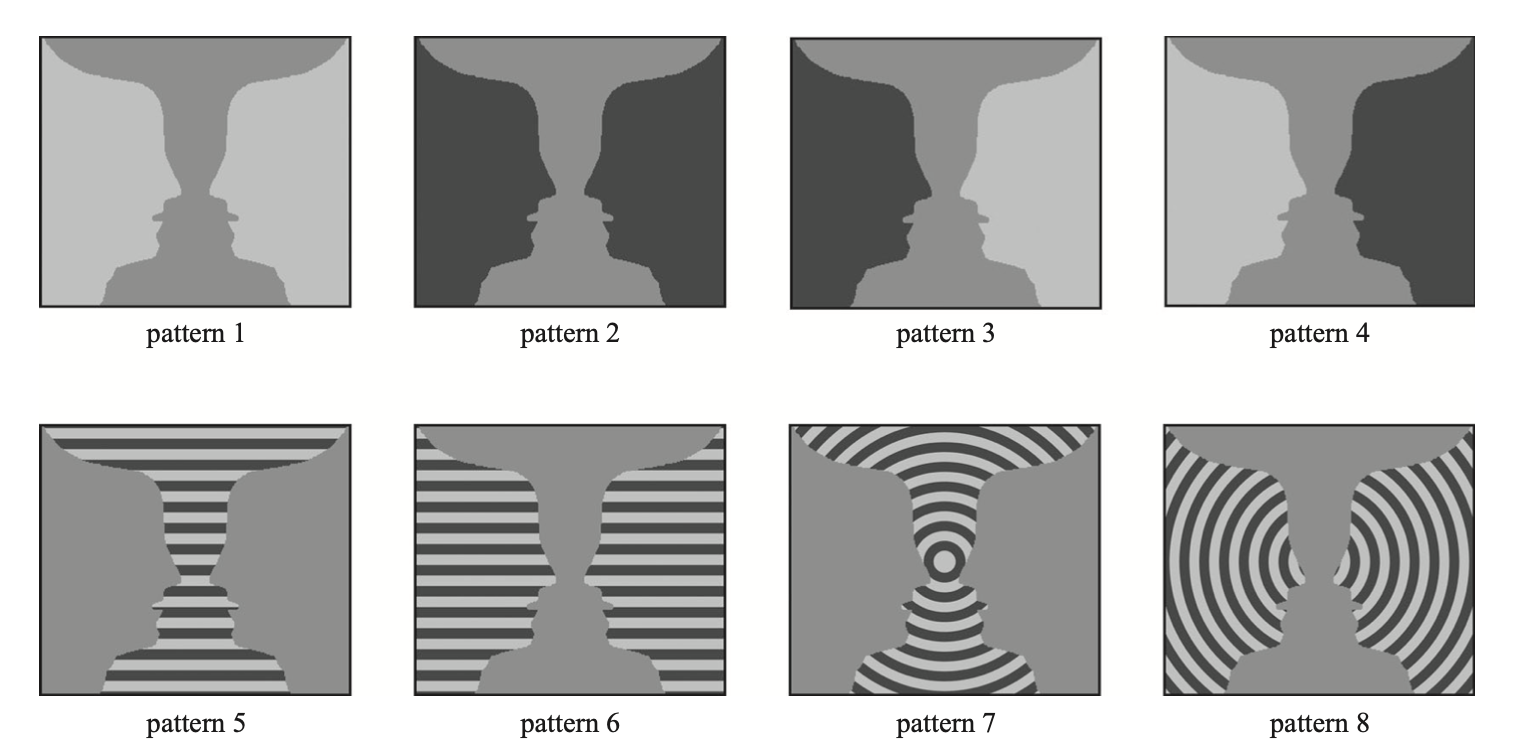}
  \caption{Variations of Rubin Vase images presented to participants in \citet{takashima2012face}.}
  \label{fig:takashima_images}
\end{figure}
\begin{figure}[ht]
\centering
  \includegraphics[width=.8\columnwidth]{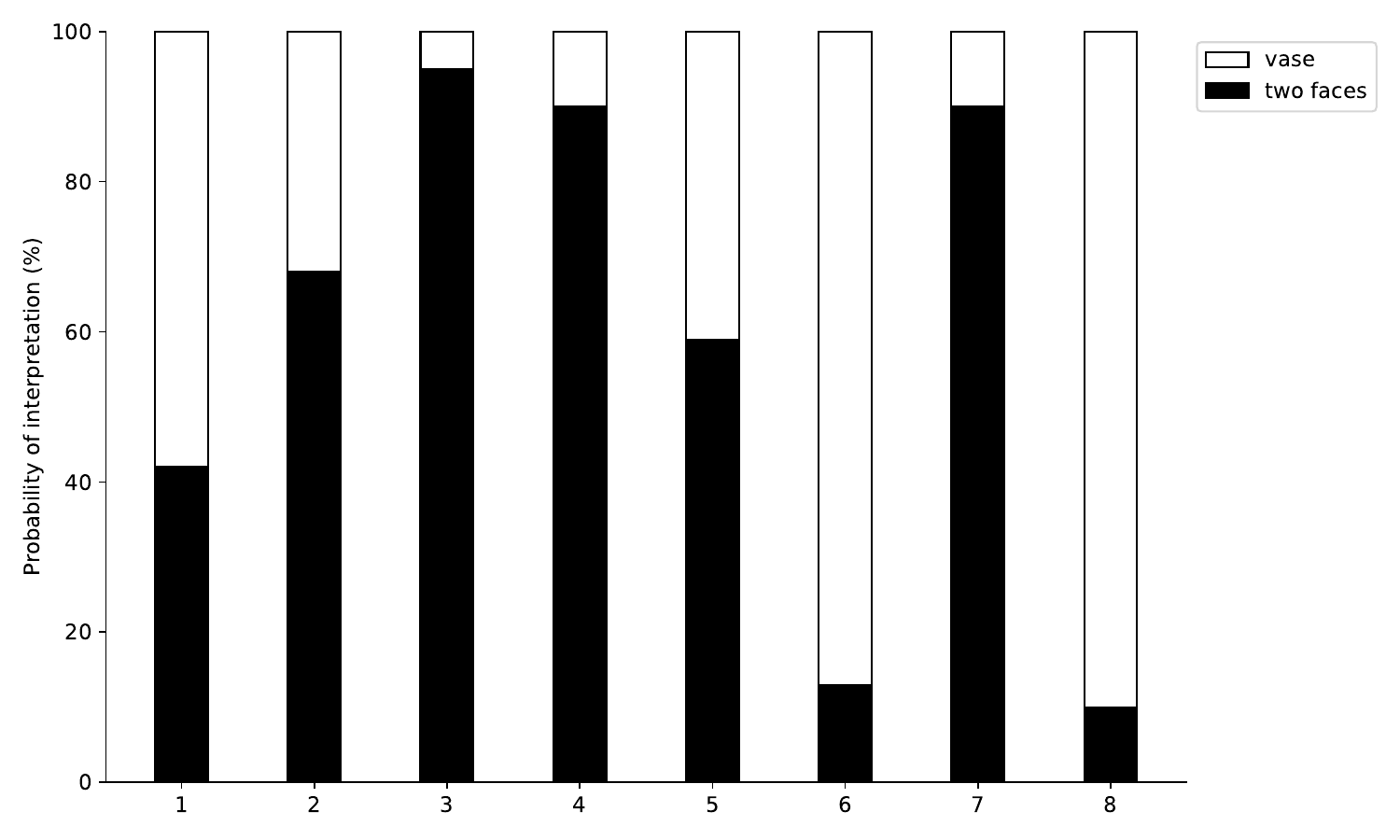}\\

  \includegraphics[width=.8\columnwidth]{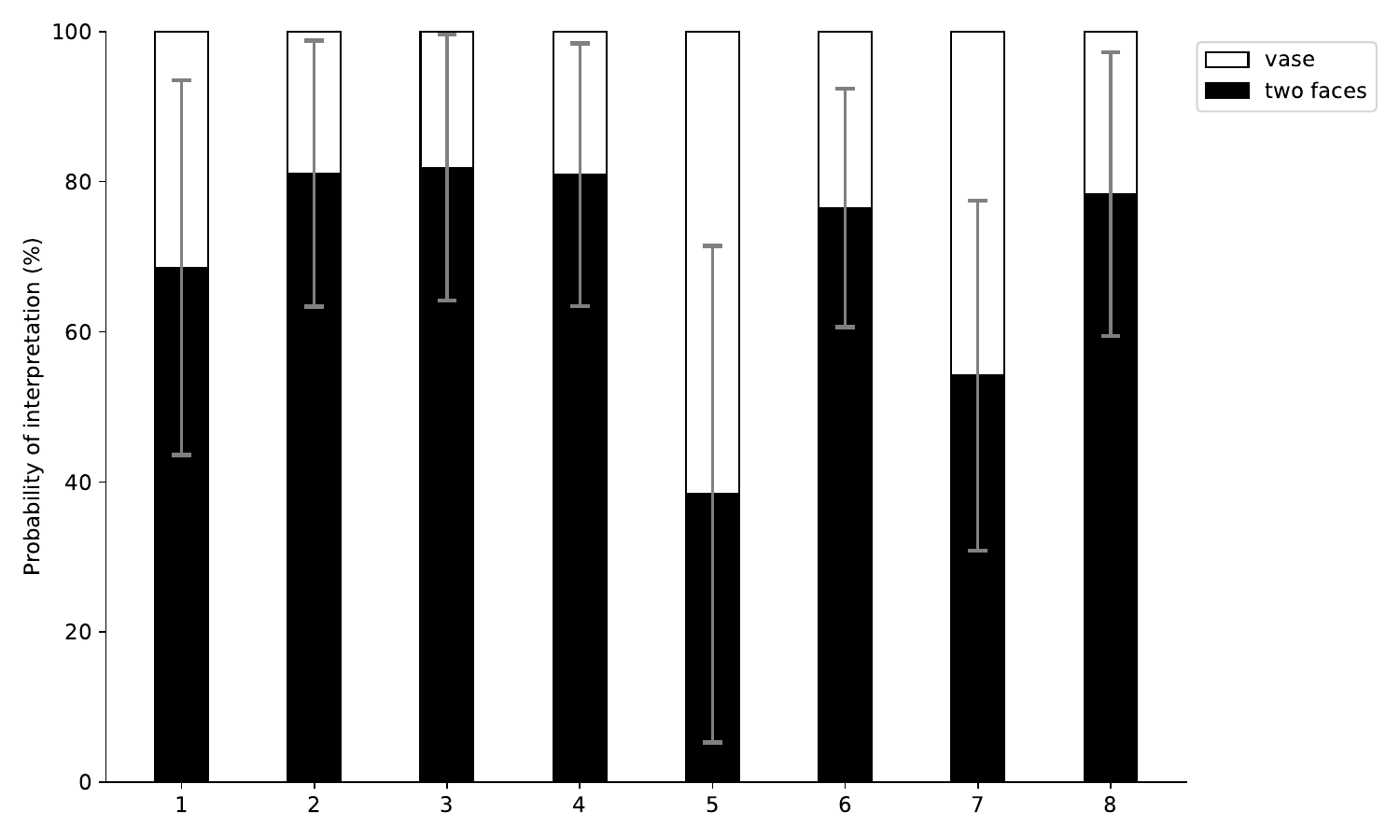}
  \caption{Comparison of between-subject average (top) and between-model average (bottom) probabilities of interpreting each image pattern as two faces in \citet{takashima2012face} our research, respectively.}
  \label{fig:takashima_plot}
\end{figure}

\subsection{Synonymous Interpretations}
To investigate the influence of synonymous interpretation labels on bistable image perception in VLMs, we substituted the original labels with synonyms. Figure \ref{fig:syn_averages} displays the effects of these changes on model preferences. The impact is generally mild, but a notable exception occurs with the 'Grimace-Begger' image, where the preference shifts dramatically. In this case, models show a clear preference for interpreting the image as a face rather than a beggar. This shift is likely attributable to the relative unfamiliarity of the synonym `panhandler' compared to the more commonly recognized term `face,' making the facial interpretation more likely for the models due to term frequency. 

\subsection{Prompt Variation}
To investigate the effect of prompt variation on VLM bistable image interpretations we examine 7 different prompts. Figure \ref{fig:prompt_averages} shows little variation on average, however, the individual decomposition of the results in figure \ref{fig:prompt} shows significant variations within models, especially for CLIP-B/32 and CLIP-B/32-Laion. In fact, while these two models are trained on distinct data of different sizes (400M vs 2B) they exhibit identical behavior across manipulations, indicating the improtance of the architecture in bistable image interpretation. The BLIP family models show significant variation in prompt compared to LLaVA and Idefics variants. This is likely due to the conditioning of the visual feature extraction module to the instruction prompt. 

\subsection{Human Interpretations}
To compare human initial interpretations with model preferences, we conducted a human evaluation using all original bistable images from our dataset, except for those from Takashima's study~\citep{takashima2012face}. We presented these images to three human annotators, asking them to identify "which interpretation they saw first?" They were also given the option to select an alternative interpretation. Figure \ref{fig:human_averages} displays the average results for each interpretation, calculated based on the frequency each interpretation was selected by the annotators across all annotations for that image.

The results reveal a limited correspondence between human and VLM interpretations, contrasting with findings for geometric illusions~\citep{afifi2019else, gomez2019convolutional, gomez2020color}. This discrepancy suggests that the training datasets for VLMs do not trigger the same cognitive biases as those encoded in humans through everyday environmental interactions and conceptual influences. It's important to note that all annotators are students at an American institution, which might influence the results; interpretations could vary significantly based on different socio-cultural experiences and the priors encoded through them.

\subsection{Replicating \citet{takashima2012face}}

We sought to evaluate VLM-human alignment on bistable image processing by comparing our results to a human study. \citet{takashima2012face} presented eight versions of the Rubin Vase illusion n=70 participants. The images are shown in Figure \ref{fig:takashima_images} and the human results are shown in the left Figure \ref{fig:takashima_plot}. They highlight two primary findings: favored 1two faces' interpretation for patterns where the profiles' homogeneity is broken (patterns 3 and 4) and favored `vase' interpretation for patterns where the faces form a continuous background by Gestalt principles~\citep{koffka1922perception} (patterns 6 and 8).

VLMs did not replicate these results, as per the right Figure \ref{fig:takashima_plot}. While the models exhibited a strong preference for the `two faces' interpretation on patterns 3 and 4, the same preference is exhibited in patterns 1 and 2 (where profiles are homogeneous). Furthermore, the models did not exhibit any preference for the `vase' interpretation in patterns 6 and 8.  Even when examined individually in Figure \ref{fig:takashima_images} (Appendix) no model exhibited similar patterns to humans. Similar to earlier results, LLaVA and Idefics variants showed high consistency across the images in their tamed preferences. The CLIP variants showed identical patterns despite the varrying patch size, unlike in the more global interventions of tint and brightness. Finally, BLIP-2 variants trained on the same image-text data with different LLMs show starkly different preferences, reinforcing the importance of language priors.

\begin{figure*}[h]
  \includegraphics[width=\linewidth]{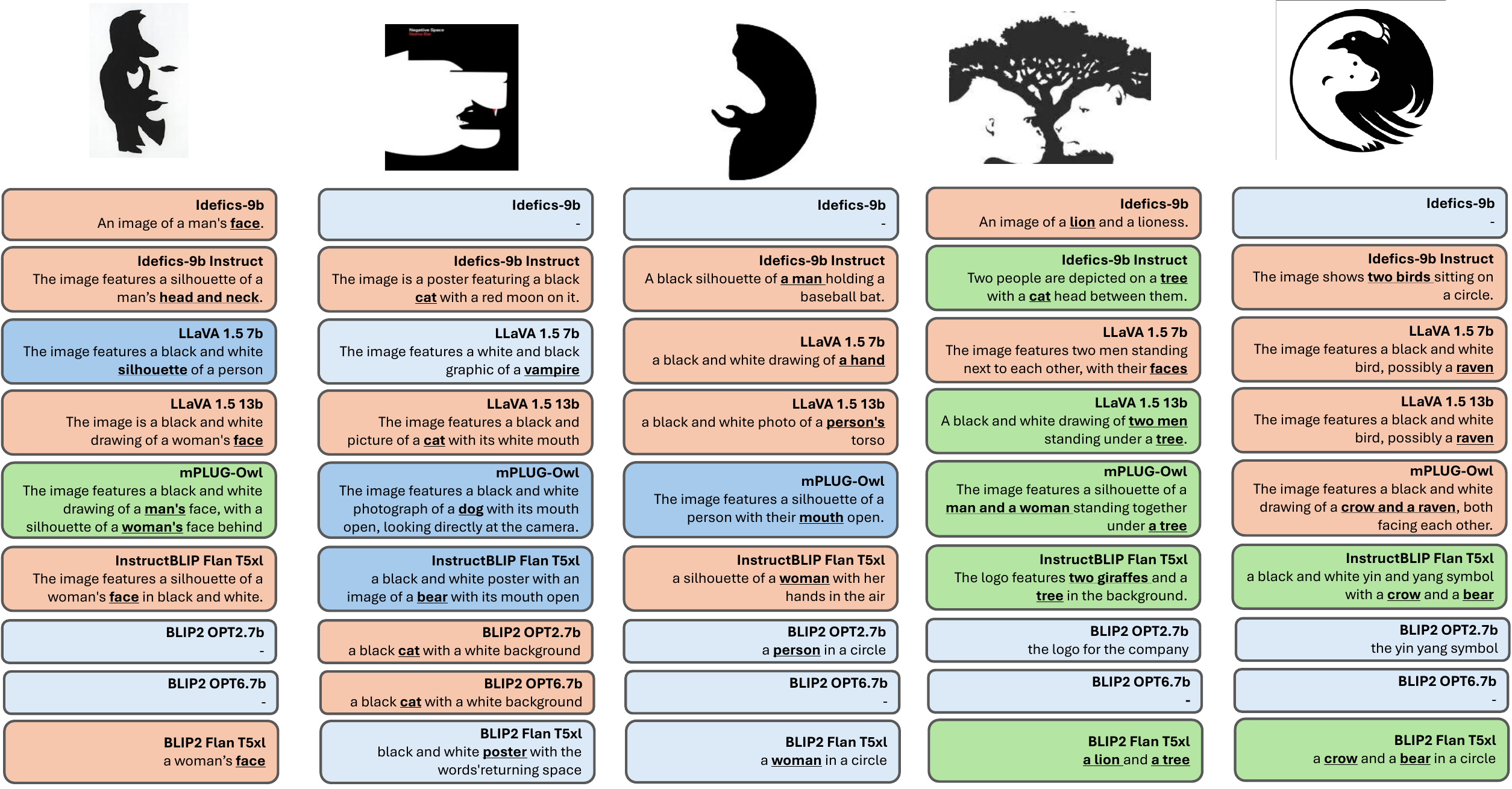}
  \caption{Depiction of generative models' descriptions for various bistable images. Orange and darker blue colors indicate selection of one of the two interpretations. Green indicate recognition of both, and light blue failure to do either.}
  \label{fig:generative_examples_many}
\end{figure*}

\subsection{Generative Results}
In the generative setup, we performed a qualitative analysis of the results. We found that several interpretation preferences discovered in the classification setup were amplified in generation. Across models, the heavily favored interpretations were faces and ducks for Rubin Vase and Duck-Rabbit images. Figure \ref{fig:generative_example} shows the output of each generative model when prompted to describe a Duck-Rabbit image. Each model employs its own explanatory style, but all favor the duck interpretation. Few models commented on the age of the individual in Young-Old Woman images, but the majority of those comments described the woman as a ``girl" or ``young woman."  An overview of the responses of the models on a subset of the images is delineated in figure \ref{fig:generative_examples_many} and all examples are listed in the Appendix Section \ref{app:qualitative}. We observe that most models only comment on a single interpretation, if at all, with some notable exceptions highlighted in green. We find that for the lion-gorilla-tree image, models are able to identify at least one of the animals, and the tree almost consistently. We hypothesize that this is because of the detail expressed in both interpretations of the image, making it easier even for humans to consciously identify both interpretations simultaneously, even if they are unable to visually perceive both at the same time. Indeed, in the human study, the `Lion-Gorilla-Tree' image received the most balanced responses across the annotators. 

\section{Discussion and Limitations}

This original analysis of VLM behavior on bistable images has yielded some interesting preliminary results. Similar to humans, VLMs have preferred initial interpretations for most classical bistable images. Five out of six models showed a preference for 'two faces' in Rubin Vase images, 'a cube seen from above' in Necker Cube images, and 'a duck' in Duck-Rabbit images. Young-Old Woman images were the only category where models' preferences were more neutral and mixed.   

We have seen minimal alignment between VLMs and humans when replicating \citet{takashima2012face} and conducting human annotations on the rest of the images. This analysis highlights that VLMs are not sensitive to the same variations that heavily impact human preferences. Models vary greatly in their sensitivity to bistablility. CLIP emerged as a model with strong, variable preferences, while LLaVa is more neutral. 

Nevertheless, making comparisons between human and machine perception of bistable images is difficult beyond the initial biases. Human perception of bistable images exhibits the phenomenon of switching interpretations through extended focus on the image. Replicating the phenomenon of switching is difficult because VLMs take static images at a single point in time. We loosely approximated the movement of time by testing the models on dozens of subtle variations of each image, as discussed above. Under the theories that subtle bottom-up cues precipitate switching in human processing, VLMs do not replicate this phenomenon. We saw that all models' preferences remained steady with variations in brightness, color, and color tint intensity. Nevertheless, this was in contrast to linguistic variations highlighting the importance of language priors in generative VLMs. 

More research is needed to further our understanding of VLM bistable image interpretation. Using VLMs that process videos could be a tractable way of mimicking the passage of time. Furthermore, additional interventions through design manipulations either through the employment of text-to-image models or human artists could reveal additional insight on VLM behavior for bistable image inputs. 

\section{Conclusion}
In this study we explore the behavior of VLMs on bistable images. We construct the largest bistable image dataset and evaluate 12 different models across six model families under various perturbations: pixel-color based perturbations, rotations, interpretation label synonyms, and prompt variations. We find that prompts have the highest impact on model preferences whereas, pixel-color perturbations have minimal effects. We further conduct human study comparisons, and find that VLMs do not exhibit the same initial biases on bistable images as human subjects. 

\section*{Acknowledgments}
This research was supported by a gift from AWS AI for research in Trustworthy AI.

\bibliography{custom}

\begin{thebibliography}{60}
\providecommand{\natexlab}[1]{#1}

\bibitem[{Afifi and Brown(2019)}]{afifi2019else}
Mahmoud Afifi and Michael~S Brown. 2019.
\newblock What else can fool deep learning? addressing color constancy errors on deep neural network performance.
\newblock In \emph{Proceedings of the IEEE/CVF international conference on computer vision}, pages 243--252.

\bibitem[{Benjamin et~al.(2019)Benjamin, Qiu, Zhang, Kording, and Stocker}]{benjamin2019shared}
Ari Benjamin, Cheng Qiu, Ling-Qi Zhang, Konrad Kording, and Alan Stocker. 2019.
\newblock Shared visual illusions between humans and artificial neural networks.
\newblock In \emph{2019 Conference on Cognitive Computational Neuroscience}, volume~10, pages 2019--1299. Cognitive Computational Neuroscience.

\bibitem[{Carlini et~al.(2023)Carlini, Jagielski, Choquette-Choo, Paleka, Pearce, Anderson, Terzis, Thomas, and Tram{\`e}r}]{carlini2023poisoning}
Nicholas Carlini, Matthew Jagielski, Christopher~A Choquette-Choo, Daniel Paleka, Will Pearce, Hyrum Anderson, Andreas Terzis, Kurt Thomas, and Florian Tram{\`e}r. 2023.
\newblock Poisoning web-scale training datasets is practical.
\newblock \emph{arXiv preprint arXiv:2302.10149}.

\bibitem[{Changpinyo et~al.(2021)Changpinyo, Sharma, Ding, and Soricut}]{changpinyo2021conceptual}
Soravit Changpinyo, Piyush Sharma, Nan Ding, and Radu Soricut. 2021.
\newblock Conceptual 12m: Pushing web-scale image-text pre-training to recognize long-tail visual concepts.
\newblock In \emph{Proceedings of the IEEE/CVF Conference on Computer Vision and Pattern Recognition}, pages 3558--3568.

\bibitem[{Chen et~al.(2015)Chen, Fang, Lin, Vedantam, Gupta, Doll{\'a}r, and Zitnick}]{chen2015microsoft}
Xinlei Chen, Hao Fang, Tsung-Yi Lin, Ramakrishna Vedantam, Saurabh Gupta, Piotr Doll{\'a}r, and C~Lawrence Zitnick. 2015.
\newblock Microsoft coco captions: Data collection and evaluation server.
\newblock \emph{arXiv preprint arXiv:1504.00325}.

\bibitem[{Chiang et~al.(2023)Chiang, Li, Lin, Sheng, Wu, Zhang, Zheng, Zhuang, Zhuang, Gonzalez, Stoica, and Xing}]{vicuna2023}
Wei-Lin Chiang, Zhuohan Li, Zi~Lin, Ying Sheng, Zhanghao Wu, Hao Zhang, Lianmin Zheng, Siyuan Zhuang, Yonghao Zhuang, Joseph~E. Gonzalez, Ion Stoica, and Eric~P. Xing. 2023.
\newblock \href {https://lmsys.org/blog/2023-03-30-vicuna/} {Vicuna: An open-source chatbot impressing gpt-4 with 90\%* chatgpt quality}.

\bibitem[{Chung et~al.(2024)Chung, Hou, Longpre, Zoph, Tay, Fedus, Li, Wang, Dehghani, Brahma et~al.}]{chung2024scaling}
Hyung~Won Chung, Le~Hou, Shayne Longpre, Barret Zoph, Yi~Tay, William Fedus, Yunxuan Li, Xuezhi Wang, Mostafa Dehghani, Siddhartha Brahma, et~al. 2024.
\newblock Scaling instruction-finetuned language models.
\newblock \emph{Journal of Machine Learning Research}, 25(70):1--53.

\bibitem[{Dai et~al.(2023)Dai, Li, Li, Tiong, Zhao, Wang, Li, Fung, and Hoi}]{dai2023instructblip}
Wenliang Dai, Junnan Li, Dongxu Li, Anthony Tiong, Junqi Zhao, Weisheng Wang, Boyang Li, Pascale Fung, and Steven Hoi. 2023.
\newblock \href {https://openreview.net/forum?id=vvoWPYqZJA} {Instruct{BLIP}: Towards general-purpose vision-language models with instruction tuning}.
\newblock In \emph{Thirty-seventh Conference on Neural Information Processing Systems}.

\bibitem[{Desai et~al.(2021)Desai, Kaul, Aysola, and Johnson}]{desai2021redcaps}
Karan Desai, Gaurav Kaul, Zubin Aysola, and Justin Johnson. 2021.
\newblock \href {https://arxiv.org/abs/2111.11431} {Redcaps: web-curated image-text data created by the people, for the people}.
\newblock \emph{Preprint}, arXiv:2111.11431.

\bibitem[{Ding et~al.(2023)Ding, Chen, Xu, Qin, Hu, Liu, Sun, and Zhou}]{ding2023enhancing}
Ning Ding, Yulin Chen, Bokai Xu, Yujia Qin, Shengding Hu, Zhiyuan Liu, Maosong Sun, and Bowen Zhou. 2023.
\newblock Enhancing chat language models by scaling high-quality instructional conversations.
\newblock In \emph{The 2023 Conference on Empirical Methods in Natural Language Processing}.

\bibitem[{Gadre et~al.(2024)Gadre, Ilharco, Fang, Hayase, Smyrnis, Nguyen, Marten, Wortsman, Ghosh, Zhang et~al.}]{gadre2024datacomp}
Samir~Yitzhak Gadre, Gabriel Ilharco, Alex Fang, Jonathan Hayase, Georgios Smyrnis, Thao Nguyen, Ryan Marten, Mitchell Wortsman, Dhruba Ghosh, Jieyu Zhang, et~al. 2024.
\newblock Datacomp: In search of the next generation of multimodal datasets.
\newblock \emph{Advances in Neural Information Processing Systems}, 36.

\bibitem[{Gomez-Villa et~al.(2019)Gomez-Villa, Martin, Vazquez-Corral, and Bertalm{\'\i}o}]{gomez2019convolutional}
Alexander Gomez-Villa, Adrian Martin, Javier Vazquez-Corral, and Marcelo Bertalm{\'\i}o. 2019.
\newblock Convolutional neural networks can be deceived by visual illusions.
\newblock In \emph{Proceedings of the IEEE/CVF conference on computer vision and pattern recognition}, pages 12309--12317.

\bibitem[{Gomez-Villa et~al.(2020)Gomez-Villa, Mart{\'\i}n, Vazquez-Corral, Bertalm{\'\i}o, and Malo}]{gomez2020color}
Alexander Gomez-Villa, Adrian Mart{\'\i}n, Javier Vazquez-Corral, Marcelo Bertalm{\'\i}o, and Jes{\'u}s Malo. 2020.
\newblock Color illusions also deceive cnns for low-level vision tasks: Analysis and implications.
\newblock \emph{Vision Research}, 176:156--174.

\bibitem[{Goyal et~al.(2017)Goyal, Khot, Summers{-}Stay, Batra, and Parikh}]{balanced_vqa_v2}
Yash Goyal, Tejas Khot, Douglas Summers{-}Stay, Dhruv Batra, and Devi Parikh. 2017.
\newblock Making the {V} in {VQA} matter: Elevating the role of image understanding in {V}isual {Q}uestion {A}nswering.
\newblock In \emph{Conference on Computer Vision and Pattern Recognition (CVPR)}.

\bibitem[{Guan et~al.(2023)Guan, Liu, Wu, Xian, Li, Liu, Wang, Chen, Huang, Yacoob et~al.}]{guan2023hallusionbench}
Tianrui Guan, Fuxiao Liu, Xiyang Wu, Ruiqi Xian, Zongxia Li, Xiaoyu Liu, Xijun Wang, Lichang Chen, Furong Huang, Yaser Yacoob, et~al. 2023.
\newblock Hallusionbench: An advanced diagnostic suite for entangled language hallucination \& visual illusion in large vision-language models.
\newblock \emph{arXiv preprint arXiv:2310.14566}.

\bibitem[{Hudson and Manning(2019)}]{hudson2019gqa}
Drew~A Hudson and Christopher~D Manning. 2019.
\newblock Gqa: A new dataset for real-world visual reasoning and compositional question answering.
\newblock In \emph{Proceedings of the IEEE/CVF conference on computer vision and pattern recognition}, pages 6700--6709.

\bibitem[{Hugrass and Crewther(2012)}]{hugrass2012willpower}
Laila Hugrass and David Crewther. 2012.
\newblock Willpower and conscious percept: volitional switching in binocular rivalry.
\newblock \emph{PloS one}, 7(4):e35963.

\bibitem[{Jaegle et~al.(2021)Jaegle, Gimeno, Brock, Vinyals, Zisserman, and Carreira}]{jaegle2021perceiver}
Andrew Jaegle, Felix Gimeno, Andy Brock, Oriol Vinyals, Andrew Zisserman, and Joao Carreira. 2021.
\newblock Perceiver: General perception with iterative attention.
\newblock In \emph{International conference on machine learning}, pages 4651--4664. PMLR.

\bibitem[{Khalil(2021)}]{khalil2021does}
Elias~L Khalil. 2021.
\newblock Why does rubin's vase differ radically from optical illusions? framing effects contra cognitive illusions.
\newblock \emph{Frontiers in Psychology}, 12:597758.

\bibitem[{Koffka(1922)}]{koffka1922perception}
Kurt Koffka. 1922.
\newblock Perception: an introduction to the gestalt-theorie.
\newblock \emph{Psychological bulletin}, 19(10):531.

\bibitem[{Kornmeier and Bach(2005)}]{KORNMEIER2005955}
Jürgen Kornmeier and Michael Bach. 2005.
\newblock \href {https://doi.org/10.1016/j.visres.2004.10.006} {The necker cube—an ambiguous figure disambiguated in early visual processing}.
\newblock \emph{Vision Research}, 45(8):955--960.

\bibitem[{Krishna et~al.(2017)Krishna, Zhu, Groth, Johnson, Hata, Kravitz, Chen, Kalantidis, Li, Shamma et~al.}]{krishna2017visual}
Ranjay Krishna, Yuke Zhu, Oliver Groth, Justin Johnson, Kenji Hata, Joshua Kravitz, Stephanie Chen, Yannis Kalantidis, Li-Jia Li, David~A Shamma, et~al. 2017.
\newblock Visual genome: Connecting language and vision using crowdsourced dense image annotations.
\newblock \emph{International journal of computer vision}, 123(1):32--73.

\bibitem[{Kuc et~al.(2023)Kuc, Maksimenko, Savosenkov, Grigorev, Grubov, Badarin, Kazantsev, Gordleeva, and Hramov}]{kuc2023studying}
Alexander Kuc, Vladimir Maksimenko, Andrey Savosenkov, Nikita Grigorev, Vadim Grubov, Artem Badarin, Victor Kazantsev, Susanna Gordleeva, and Alexander Hramov. 2023.
\newblock Studying perceptual bias in favor of the from-above necker cube perspective in a goal-directed behavior.
\newblock \emph{Frontiers in Psychology}, 14:1160605.

\bibitem[{Laing and Chow(2002)}]{laing2002spiking}
Carlo~R Laing and Carson~C Chow. 2002.
\newblock A spiking neuron model for binocular rivalry.
\newblock \emph{Journal of computational neuroscience}, 12:39--53.

\bibitem[{Lauren{\c{c}}on et~al.(2024)Lauren{\c{c}}on, Saulnier, Tronchon, Bekman, Singh, Lozhkov, Wang, Karamcheti, Rush, Kiela et~al.}]{laurenccon2024obelics}
Hugo Lauren{\c{c}}on, Lucile Saulnier, L{\'e}o Tronchon, Stas Bekman, Amanpreet Singh, Anton Lozhkov, Thomas Wang, Siddharth Karamcheti, Alexander Rush, Douwe Kiela, et~al. 2024.
\newblock Obelics: An open web-scale filtered dataset of interleaved image-text documents.
\newblock \emph{Advances in Neural Information Processing Systems}, 36.

\bibitem[{Li et~al.(2023{\natexlab{a}})Li, Li, Savarese, and Hoi}]{li2023blip}
Junnan Li, Dongxu Li, Silvio Savarese, and Steven Hoi. 2023{\natexlab{a}}.
\newblock Blip-2: Bootstrapping language-image pre-training with frozen image encoders and large language models.
\newblock In \emph{International conference on machine learning}, pages 19730--19742. PMLR.

\bibitem[{Li et~al.(2023{\natexlab{b}})Li, Li, Savarese, and Hoi}]{pmlr-v202-li23q}
Junnan Li, Dongxu Li, Silvio Savarese, and Steven Hoi. 2023{\natexlab{b}}.
\newblock \href {https://proceedings.mlr.press/v202/li23q.html} {{BLIP}-2: Bootstrapping language-image pre-training with frozen image encoders and large language models}.
\newblock In \emph{Proceedings of the 40th International Conference on Machine Learning}, volume 202 of \emph{Proceedings of Machine Learning Research}, pages 19730--19742. PMLR.

\bibitem[{Li et~al.(2021)Li, Selvaraju, Gotmare, Joty, Xiong, and Hoi}]{li2021align}
Junnan Li, Ramprasaath Selvaraju, Akhilesh Gotmare, Shafiq Joty, Caiming Xiong, and Steven Chu~Hong Hoi. 2021.
\newblock Align before fuse: Vision and language representation learning with momentum distillation.
\newblock \emph{Advances in neural information processing systems}, 34:9694--9705.

\bibitem[{Li et~al.(2023{\natexlab{c}})Li, Yin, Li, Chen, Wang, Ren, Li, Yang, Xu, Sun et~al.}]{li2023m3it}
Lei Li, Yuwei Yin, Shicheng Li, Liang Chen, Peiyi Wang, Shuhuai Ren, Mukai Li, Yazheng Yang, Jingjing Xu, Xu~Sun, et~al. 2023{\natexlab{c}}.
\newblock M3it: A large-scale dataset towards multi-modal multilingual instruction tuning.
\newblock \emph{arXiv preprint arXiv:2306.04387}.

\bibitem[{Lian et~al.(2023)Lian, Wang, Goodson, Pentland, Cook, Vong, and "Teknium"}]{SlimOrca}
Wing Lian, Guan Wang, Bleys Goodson, Eugene Pentland, Austin Cook, Chanvichet Vong, and "Teknium". 2023.
\newblock \href {https://https://huggingface.co/Open-Orca/SlimOrca} {Slimorca: An open dataset of gpt-4 augmented flan reasoning traces, with verification}.

\bibitem[{Liu et~al.(2023{\natexlab{a}})Liu, Li, Li, and Lee}]{liu2023improved}
Haotian Liu, Chunyuan Li, Yuheng Li, and Yong~Jae Lee. 2023{\natexlab{a}}.
\newblock Improved baselines with visual instruction tuning.
\newblock In \emph{NeurIPS 2023 Workshop on Instruction Tuning and Instruction Following}.

\bibitem[{Liu et~al.(2023{\natexlab{b}})Liu, Li, Wu, and Lee}]{liu2023visual}
Haotian Liu, Chunyuan Li, Qingyang Wu, and Yong~Jae Lee. 2023{\natexlab{b}}.
\newblock \href {https://openreview.net/forum?id=w0H2xGHlkw} {Visual instruction tuning}.
\newblock In \emph{Thirty-seventh Conference on Neural Information Processing Systems}.

\bibitem[{Marino et~al.(2019)Marino, Rastegari, Farhadi, and Mottaghi}]{okvqa}
Kenneth Marino, Mohammad Rastegari, Ali Farhadi, and Roozbeh Mottaghi. 2019.
\newblock Ok-vqa: A visual question answering benchmark requiring external knowledge.
\newblock In \emph{Conference on Computer Vision and Pattern Recognition (CVPR)}.

\bibitem[{Mishra et~al.(2019)Mishra, Shekhar, Singh, and Chakraborty}]{mishraICDAR19}
Anand Mishra, Shashank Shekhar, Ajeet~Kumar Singh, and Anirban Chakraborty. 2019.
\newblock Ocr-vqa: Visual question answering by reading text in images.
\newblock In \emph{ICDAR}.

\bibitem[{Moreno-Bote et~al.(2007)Moreno-Bote, Rinzel, and Rubin}]{moreno2007noise}
Rub{\'e}n Moreno-Bote, John Rinzel, and Nava Rubin. 2007.
\newblock Noise-induced alternations in an attractor network model of perceptual bistability.
\newblock \emph{Journal of neurophysiology}, 98(3):1125--1139.

\bibitem[{Ordonez et~al.(2011)Ordonez, Kulkarni, and Berg}]{ordonez2011sbucaptions}
Vicente Ordonez, Girish Kulkarni, and Tamara~L. Berg. 2011.
\newblock Im2text: Describing images using 1 million captioned photographs.
\newblock In \emph{Neural Information Processing Systems ({NIPS})}.

\bibitem[{Pont-Tuset et~al.(2020)Pont-Tuset, Uijlings, Changpinyo, Soricut, and Ferrari}]{ponttuset2020localized}
Jordi Pont-Tuset, Jasper Uijlings, Soravit Changpinyo, Radu Soricut, and Vittorio Ferrari. 2020.
\newblock Connecting vision and language with localized narratives.
\newblock In \emph{ECCV}.

\bibitem[{Radford et~al.(2021)Radford, Kim, Hallacy, Ramesh, Goh, Agarwal, Sastry, Askell, Mishkin, Clark et~al.}]{radford2021learning}
Alec Radford, Jong~Wook Kim, Chris Hallacy, Aditya Ramesh, Gabriel Goh, Sandhini Agarwal, Girish Sastry, Amanda Askell, Pamela Mishkin, Jack Clark, et~al. 2021.
\newblock Learning transferable visual models from natural language supervision.
\newblock In \emph{International conference on machine learning}, pages 8748--8763. PMLR.

\bibitem[{Schuhmann et~al.(2022)Schuhmann, Beaumont, Vencu, Gordon, Wightman, Cherti, Coombes, Katta, Mullis, Wortsman et~al.}]{schuhmann2022laion}
Christoph Schuhmann, Romain Beaumont, Richard Vencu, Cade Gordon, Ross Wightman, Mehdi Cherti, Theo Coombes, Aarush Katta, Clayton Mullis, Mitchell Wortsman, et~al. 2022.
\newblock Laion-5b: An open large-scale dataset for training next generation image-text models.
\newblock \emph{Advances in Neural Information Processing Systems}, 35:25278--25294.

\bibitem[{Schuhmann et~al.(2021)Schuhmann, Vencu, Beaumont, Kaczmarczyk, Mullis, Katta, Coombes, Jitsev, and Komatsuzaki}]{schuhmann2021laion}
Christoph Schuhmann, Richard Vencu, Romain Beaumont, Robert Kaczmarczyk, Clayton Mullis, Aarush Katta, Theo Coombes, Jenia Jitsev, and Aran Komatsuzaki. 2021.
\newblock Laion-400m: Open dataset of clip-filtered 400 million image-text pairs.
\newblock \emph{arXiv preprint arXiv:2111.02114}.

\bibitem[{Schwenk et~al.(2022)Schwenk, Khandelwal, Clark, Marino, and Mottaghi}]{AOKVQA}
Dustin Schwenk, Apoorv Khandelwal, Christopher Clark, Kenneth Marino, and Roozbeh Mottaghi. 2022.
\newblock A-okvqa: A benchmark for visual question answering using world knowledge.
\newblock In \emph{European Conference on Computer Vision}, pages 146--162. Springer.

\bibitem[{Sharma et~al.(2018)Sharma, Ding, Goodman, and Soricut}]{sharma2018conceptual}
Piyush Sharma, Nan Ding, Sebastian Goodman, and Radu Soricut. 2018.
\newblock Conceptual captions: A cleaned, hypernymed, image alt-text dataset for automatic image captioning.
\newblock In \emph{Proceedings of the 56th Annual Meeting of the Association for Computational Linguistics (Volume 1: Long Papers)}, pages 2556--2565.

\bibitem[{Sidorov et~al.(2020)Sidorov, Hu, Rohrbach, and Singh}]{sidorov2020textcaps}
Oleksii Sidorov, Ronghang Hu, Marcus Rohrbach, and Amanpreet Singh. 2020.
\newblock Textcaps: a dataset for image captioning with reading comprehension.
\newblock In \emph{Computer Vision--ECCV 2020: 16th European Conference, Glasgow, UK, August 23--28, 2020, Proceedings, Part II 16}, pages 742--758. Springer.

\bibitem[{Slotnick and Yantis(2005)}]{slotnick2005common}
Scott~D Slotnick and Steven Yantis. 2005.
\newblock Common neural substrates for the control and effects of visual attention and perceptual bistability.
\newblock \emph{Cognitive Brain Research}, 24(1):97--108.

\bibitem[{Srinivasan et~al.(2021)Srinivasan, Raman, Chen, Bendersky, and Najork}]{srinivasan2021wit}
Krishna Srinivasan, Karthik Raman, Jiecao Chen, Michael Bendersky, and Marc Najork. 2021.
\newblock Wit: Wikipedia-based image text dataset for multimodal multilingual machine learning.
\newblock \emph{arXiv preprint arXiv:2103.01913}.

\bibitem[{Sun and Dekel(2021)}]{sun2021imagenet}
Eric~D Sun and Ron Dekel. 2021.
\newblock Imagenet-trained deep neural networks exhibit illusion-like response to the scintillating grid.
\newblock \emph{Journal of Vision}, 21(11):15--15.

\bibitem[{Sun et~al.(2023)Sun, Fang, Wu, Wang, and Cao}]{sun2023eva}
Quan Sun, Yuxin Fang, Ledell Wu, Xinlong Wang, and Yue Cao. 2023.
\newblock Eva-clip: Improved training techniques for clip at scale.
\newblock \emph{arXiv preprint arXiv:2303.15389}.

\bibitem[{Takashima et~al.(2012)Takashima, Fujii, and Shiina}]{takashima2012face}
Midori Takashima, Teruo Fujii, and Ken Shiina. 2012.
\newblock Face or vase? areal homogeneity effect.
\newblock \emph{Perception}, 41(11):1392--1394.

\bibitem[{Thomee et~al.(2016)Thomee, Shamma, Friedland, Elizalde, Ni, Poland, Borth, and Li}]{thomee2016yfcc100m}
Bart Thomee, David~A Shamma, Gerald Friedland, Benjamin Elizalde, Karl Ni, Douglas Poland, Damian Borth, and Li-Jia Li. 2016.
\newblock Yfcc100m: The new data in multimedia research.
\newblock \emph{Communications of the ACM}, 59(2):64--73.

\bibitem[{Touvron et~al.(2023)Touvron, Lavril, Izacard, Martinet, Lachaux, Lacroix, Rozi{\`e}re, Goyal, Hambro, Azhar et~al.}]{touvron2023llama}
Hugo Touvron, Thibaut Lavril, Gautier Izacard, Xavier Martinet, Marie-Anne Lachaux, Timoth{\'e}e Lacroix, Baptiste Rozi{\`e}re, Naman Goyal, Eric Hambro, Faisal Azhar, et~al. 2023.
\newblock Llama: Open and efficient foundation language models.
\newblock \emph{arXiv preprint arXiv:2302.13971}.

\bibitem[{Wang et~al.(2013)Wang, Arteaga, and He}]{wang2013brain}
Megan Wang, Daniel Arteaga, and Biyu~J He. 2013.
\newblock Brain mechanisms for simple perception and bistable perception.
\newblock \emph{Proceedings of the National Academy of Sciences}, 110(35):E3350--E3359.

\bibitem[{Wang et~al.(2024)Wang, Chen, Han, Lin, Zhao, Liu, Zhai, Yuan, You, and Yang}]{wang2024exploring}
Yiqi Wang, Wentao Chen, Xiaotian Han, Xudong Lin, Haiteng Zhao, Yongfei Liu, Bohan Zhai, Jianbo Yuan, Quanzeng You, and Hongxia Yang. 2024.
\newblock Exploring the reasoning abilities of multimodal large language models (mllms): A comprehensive survey on emerging trends in multimodal reasoning.
\newblock \emph{arXiv preprint arXiv:2401.06805}.

\bibitem[{Ward and Scholl(2015)}]{ward2015stochastic}
Emily~J Ward and Brian~J Scholl. 2015.
\newblock Stochastic or systematic? seemingly random perceptual switching in bistable events triggered by transient unconscious cues.
\newblock \emph{Journal of Experimental Psychology: Human Perception and Performance}, 41(4):929.

\bibitem[{Wei et~al.(2021)Wei, Bosma, Zhao, Guu, Yu, Lester, Du, Dai, and Le}]{wei2021finetuned}
Jason Wei, Maarten Bosma, Vincent Zhao, Kelvin Guu, Adams~Wei Yu, Brian Lester, Nan Du, Andrew~M Dai, and Quoc~V Le. 2021.
\newblock Finetuned language models are zero-shot learners.
\newblock In \emph{International Conference on Learning Representations}.

\bibitem[{Ye et~al.(2023)Ye, Xu, Xu, Ye, Yan, Zhou, Wang, Hu, Shi, Shi et~al.}]{ye2023mplug}
Qinghao Ye, Haiyang Xu, Guohai Xu, Jiabo Ye, Ming Yan, Yiyang Zhou, Junyang Wang, Anwen Hu, Pengcheng Shi, Yaya Shi, et~al. 2023.
\newblock mplug-owl: Modularization empowers large language models with multimodality.
\newblock \emph{arXiv preprint arXiv:2304.14178}.

\bibitem[{Yu et~al.(2016)Yu, Poirson, Yang, Berg, and Berg}]{yu2016modeling}
Licheng Yu, Patrick Poirson, Shan Yang, Alexander~C Berg, and Tamara~L Berg. 2016.
\newblock Modeling context in referring expressions.
\newblock In \emph{Computer Vision--ECCV 2016: 14th European Conference, Amsterdam, The Netherlands, October 11-14, 2016, Proceedings, Part II 14}, pages 69--85. Springer.

\bibitem[{Zhang et~al.(2022)Zhang, Roller, Goyal, Artetxe, Chen, Chen, Dewan, Diab, Li, Lin et~al.}]{zhang2022opt}
Susan Zhang, Stephen Roller, Naman Goyal, Mikel Artetxe, Moya Chen, Shuohui Chen, Christopher Dewan, Mona Diab, Xian Li, Xi~Victoria Lin, et~al. 2022.
\newblock Opt: Open pre-trained transformer language models.
\newblock \emph{arXiv preprint arXiv:2205.01068}.

\bibitem[{Zhang et~al.(2023{\natexlab{a}})Zhang, Zhang, Gu, Zhou, Lipka, Yang, and Sun}]{zhang2023llavar}
Yanzhe Zhang, Ruiyi Zhang, Jiuxiang Gu, Yufan Zhou, Nedim Lipka, Diyi Yang, and Tong Sun. 2023{\natexlab{a}}.
\newblock Llavar: Enhanced visual instruction tuning for text-rich image understanding.
\newblock \emph{arXiv preprint arXiv:2306.17107}.

\bibitem[{Zhang et~al.(2023{\natexlab{b}})Zhang, Pan, Zhou, Pan, and Chai}]{zhang2023grounding}
Yichi Zhang, Jiayi Pan, Yuchen Zhou, Rui Pan, and Joyce Chai. 2023{\natexlab{b}}.
\newblock \href {https://arxiv.org/abs/2311.00047} {Grounding visual illusions in language: Do vision-language models perceive illusions like humans?}
\newblock \emph{Preprint}, arXiv:2311.00047.

\bibitem[{Zhao et~al.(2023)Zhao, Wu, and Huang}]{zhao2023svit}
Bo~Zhao, Boya Wu, and Tiejun Huang. 2023.
\newblock Svit: Scaling up visual instruction tuning.
\newblock \emph{arXiv preprint arXiv:2307.04087}.

\end{thebibliography}

\appendix

\section{Model Details}
\label{app:models}
We summarize the architectural differences for the models used in our study in Table \ref{tab:models} and list the various datasets they were trained on both for pretraining and instruction tuning (where applicable) in Table \ref{tab:model_data}.
\begin{table*}[htbp]
\fontsize{4}{4}\selectfont
\centering
\begin{tabularx}{\textwidth}{lXcXcXccccXX}
\toprule
\textbf{Model} &\textbf{\#Train Param.}&  \textbf{LLM}& \textbf{Res.}&\textbf{ViT}&  \textbf{LLM Size}& \textbf{V-L ~~Type}& \textbf{V-L ~~Size} &\textbf{\#Tokens}& \textbf{Deep V-L} & \textbf{Frozen LLM} & \textbf{Frozen ViT}\\
\midrule
Idefics 9b~\citep{laurenccon2024obelics} & 9b &LLaMA\citep{touvron2023llama} & 224& OpenCLIP-H\footnote{https://laion.ai/blog/large-openclip/}&7b & Perceiver~\citep{jaegle2021perceiver} &194M & 64&\checkmark&\checkmark &  \checkmark \\

Idefics 9b Instruct &9b& LlaMA & 224& OpenCLIP&7b &Perceiver  & 194M & 64&\checkmark& \checkmark& \checkmark\\\midrule

LLaVA-1.5 7b~\citep{liu2023visual,liu2023improved} &20M&Vicuna1.5-7B~\citep{vicuna2023}&336&CLIP ViT-L~\citep{radford2021learning}&7b & Linear  & 20M&577&$\times$&$\times$&\checkmark \\
LLaVA-1.5 13b&20M &Vicuna1.5-13B&336&CLIP ViT-L& 13b& Linear  & 20M&577&$\times$ &$\times$& \checkmark \\\midrule

BLIP-2 OPT2.7b~\citep{li2023blip} &188M &OPT2.7b~\citep{zhang2022opt}&224&EVA-CLIP-g& 2.7b &Q-Former &188M &32&$\times$&\checkmark & \checkmark \\
BLIP-2 OPT6.7b &188M &OPT6.7b&224&EVA-CLIP-g&6.7b &Q-Former &188M&32&$\times$ &\checkmark &\checkmark\\
BLIP-2 FlanT5xl &188M &FlanT5xl&224&EVA-CLIP-g& 3b &Q-Former&188M &32&$\times$&\checkmark & \checkmark \\\midrule

InstructBLIP FlanT5xl~\citep{dai2023instructblip} &188M &FlanT5xl~\citep{chung2024scaling}&224&EVA-CLIP-g\citep{sun2023eva}& 3b &Q-Former~\citep{pmlr-v202-li23q} &188M &32&$\times$&\checkmark & \checkmark \\\midrule

mPLUG-Owl~\citep{ye2023mplug} & 500M &LLaMA&224&CLIP ViT-L&7b&Visual Abstractor~\citep{ye2023mplug}&&64&$\times$&$\times$&$\times$\\
\bottomrule
\end{tabularx}
    \caption{Overview of Generative VLMs architectures examined on their perception of bistable images. }
    \label{tab:models}
\end{table*}

\begin{table*}[htbp]
\fontsize{5}{5}\selectfont
\centering
\begin{tabularx}{\textwidth}{lXX}
\toprule
\textbf{Model} &\textbf{Pretraining Data}&  \textbf{Instruction Tuning Data}\\
\midrule
CLIP & 400M image-caption data (undisclosed) & N/A\\
Idefics 9b & OBELICS~\citep{laurenccon2024obelics}, Wikipedia\footnote{https://dumps.wikimedia.org},Conceptual Captions\citep{sharma2018conceptual}, Conceptual Captions 12M~\citep{changpinyo2021conceptual}, WIT~\citep{srinivasan2021wit}, Localized Narratives~\citep{ponttuset2020localized}, RedCaps~\citep{desai2021redcaps}, COCO~\citep{chen2015microsoft}, SBU Captions~\citep{ordonez2011sbucaptions}, Visual Genome~\citep{krishna2017visual}, YFCC100M~\citep{thomee2016yfcc100m} & N/A \\

Idefics 9b Instruct & OBELICS~\citep{laurenccon2024obelics}, Wikipedia\footnote{https://dumps.wikimedia.org},CC3M\citep{sharma2018conceptual}, CC12M~\citep{changpinyo2021conceptual}, WIT~\citep{srinivasan2021wit}, Localized Narratives~\citep{ponttuset2020localized}, RedCaps~\citep{desai2021redcaps}, COCO~\citep{chen2015microsoft}, SBU~\citep{ordonez2011sbucaptions}, Visual Genome~\citep{krishna2017visual}, YFCC100M~\citep{thomee2016yfcc100m}& M3IT~\citep{li2023m3it}, LRV-Instruction~\citep{}, LLaVA150k~\citep{liu2023visual},LLaVAR-Instruct~\citep{zhang2023llavar},SVIT~\citep{zhao2023svit}, UltraChat~\citep{ding2023enhancing} \\

LLaVA-1.5 & LLaVA~\citep{liu2023improved} [subsets of LAION-400M~\citep{schuhmann2021laion}, CC3M~\citep{sharma2018conceptual}, SBU~\citep{ordonez2011sbucaptions}]& VQAv2~\citep{balanced_vqa_v2}, GQA~\citep{hudson2019gqa},OKVQA~\citep{okvqa}, A-OKVQA~\citep{AOKVQA},OCRVQA~\cite{mishraICDAR19}, TextCaps~\citep{sidorov2020textcaps}, LLaVA150k~\citep{liu2023visual}, ShareGPT~\footnote{(https://sharegpt.com/}  \\

BLIP-2 &COCO~\citep{chen2015microsoft}, CC3M~\citep{sharma2018conceptual}, CC12M~\citep{changpinyo2021conceptual}, LAION400M~\citep{schuhmann2021laion}, Visual Genome~\citep{krishna2017visual}&N/A \\

InstructBLIP &COCO~\citep{chen2015microsoft}, CC3M~\citep{sharma2018conceptual}, CC12M~\citep{changpinyo2021conceptual}, LAION400M~\citep{schuhmann2021laion}, Visual Genome~\citep{krishna2017visual}& COCO~\citep{chen2015microsoft}, Web CapFilt~\citep{li2023blip}, TextCaps~\citep{sidorov2020textcaps}, VQAv2~\citep{balanced_vqa_v2}, OKVQA~\citep{okvqa}, A-OKVQA~\citep{AOKVQA}, LLaVA150k~\citep{liu2023visual}, OCRVQA~\cite{mishraICDAR19} \\

mPLUG-Owl & LAION-400M~\citep{schuhmann2021laion}, COYO~\citep{carlini2023poisoning}, COCO~\citep{chen2015microsoft}, Laion-en~\citep{schuhmann2022laion}, DataComp~\citep{gadre2024datacomp}& VQAv2~\citep{balanced_vqa_v2}, OKVQA~\citep{okvqa}, OCR-VQA~\citep{mishraICDAR19}, GQA~\citep{hudson2019gqa}, A-OKVQA~\citep{AOKVQA}, RefCOCO~\citep{yu2016modeling}, Visual Genome~\citep{krishna2017visual}, LLaVA150K~\citep{liu2023visual}, ShareGPT, SlimOrca~\citep{SlimOrca} \\
\bottomrule
\end{tabularx}
    \caption{Overview of Pretraining and Instruction Tuning Datasets (adapted from \citet{wang2024exploring})}
    \label{tab:model_data}
\end{table*}

\section{Additional Results}
\subsection{Individual Results: Original Images}
\begin{figure*}[ht]
\includegraphics[width=\linewidth]{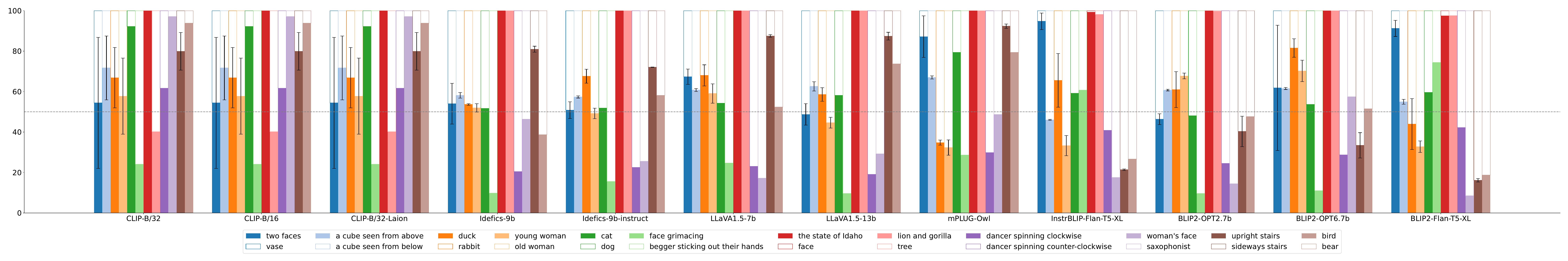} 
  \caption {Average probability distributions for each model evaluated on each image category.}
\label{fig:models_averages}
\end{figure*}

Figure \ref{fig:models_averages} we list the individual model results for the original labels.

\subsection{Individual Results: Synonymous Interpretations}
\begin{figure*}[h]
    \centering
    \includegraphics[width=\linewidth]{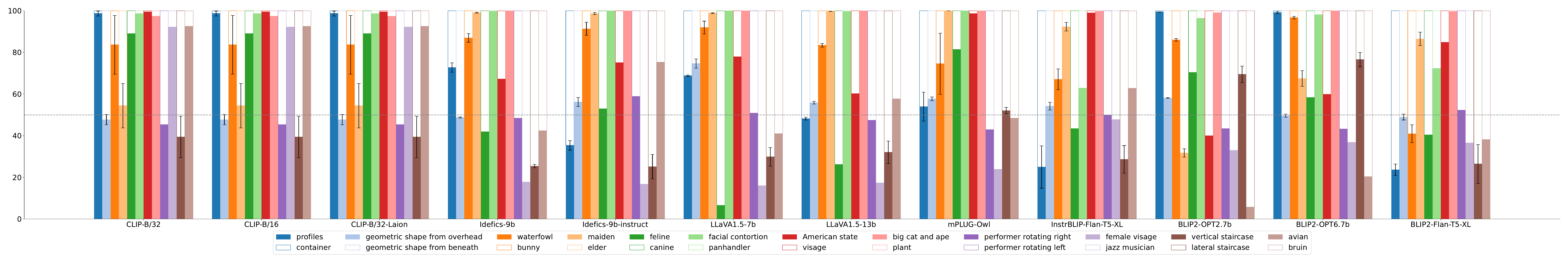}
    \caption{Synonymous Interpretations}
    \label{fig:synonymous_individual}
\end{figure*}

Figure \ref{fig:synonymous_individual} we list the individual model results for the synononymous labels. We find that there is non-trivial variation that is attributed to the likelihood of the terms used as the labels. 

\subsection{Individual Results:  \citet{takashima2012face}}
\begin{figure*}[t]
  \includegraphics[width=\linewidth]{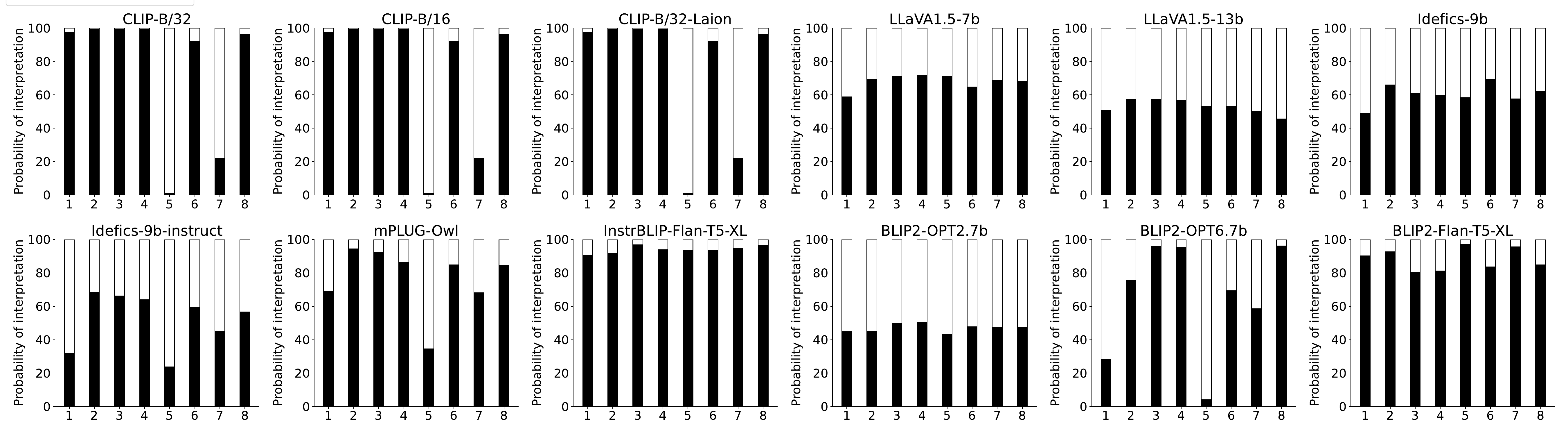}
  \caption{Individual model preferences for  \citet{takashima2012face} images.}
  \label{fig:takashima_plot_indiv}
\end{figure*}

Figure \ref{fig:takashima_plot_indiv} lists the individual results for each model on the \citet{takashima2012face} image study. 

\subsection{Tint Variation Individual Plots}
\begin{figure*}[h]
    \centering
    \includegraphics[width=\linewidth]{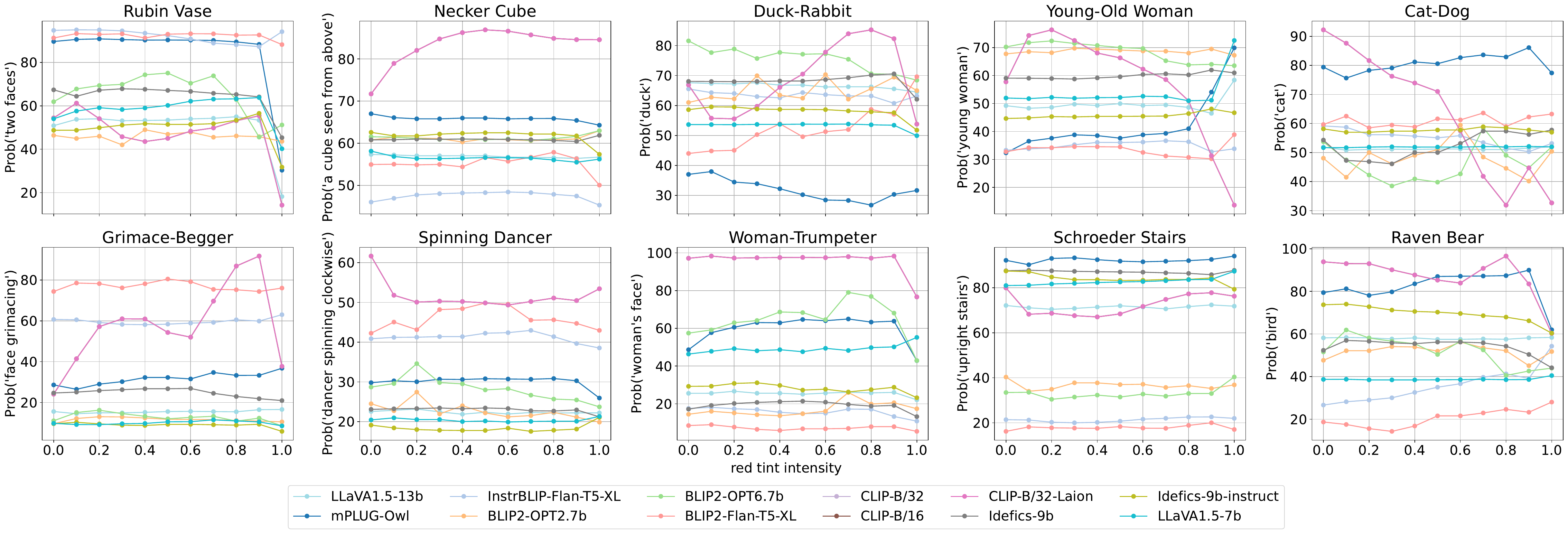}
    \caption{Red Tint Variation}
    \label{fig:red_tint}
\end{figure*}

\begin{figure*}[h]
    \centering
    \includegraphics[width=\linewidth]{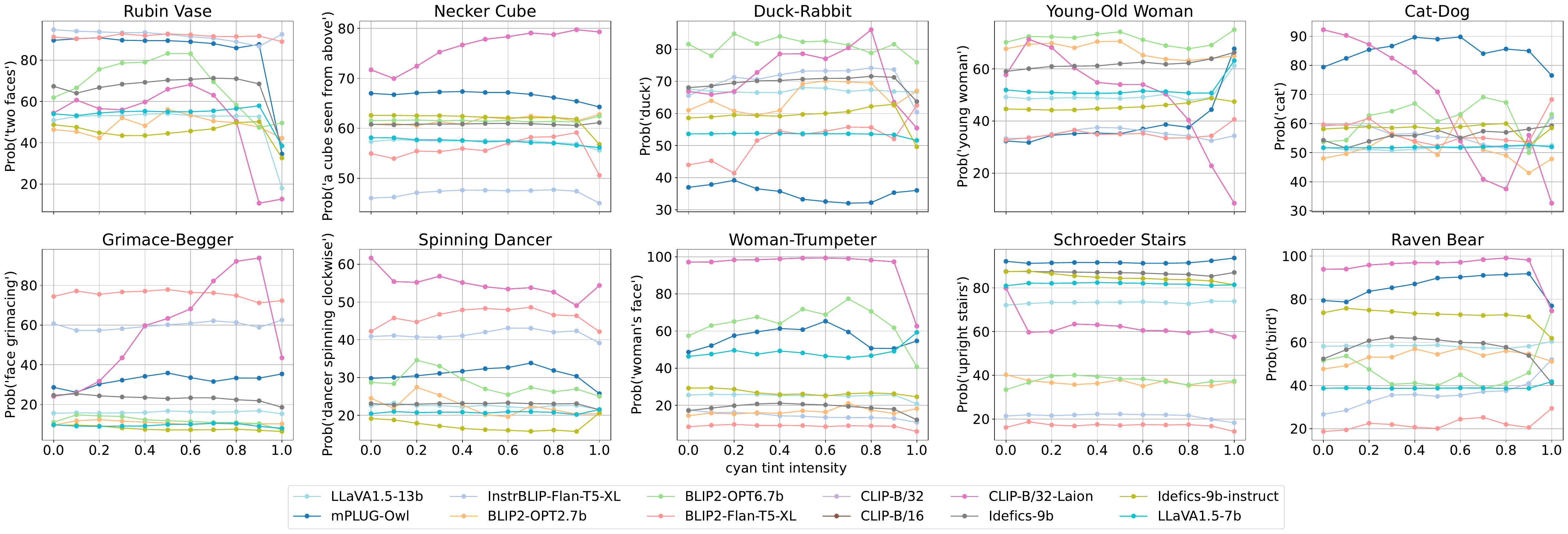}
    \caption{Cyan Tint Variation}
        \label{fig:cyan_tint}
\end{figure*}

\begin{figure*}[h]
    \centering
    \includegraphics[width=\linewidth]{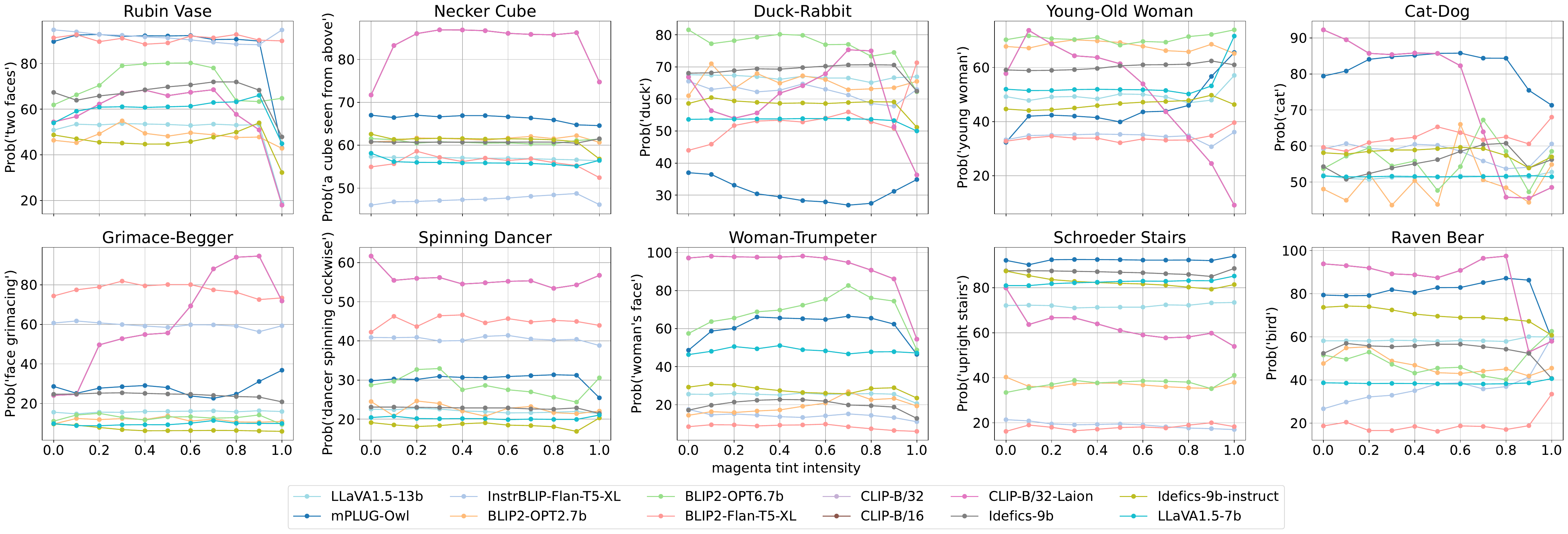}
    \caption{Magenta Tint Variation}
     \label{fig:magenta_tint}
\end{figure*}

\begin{figure*}[h]
    \centering
    \includegraphics[width=\linewidth]{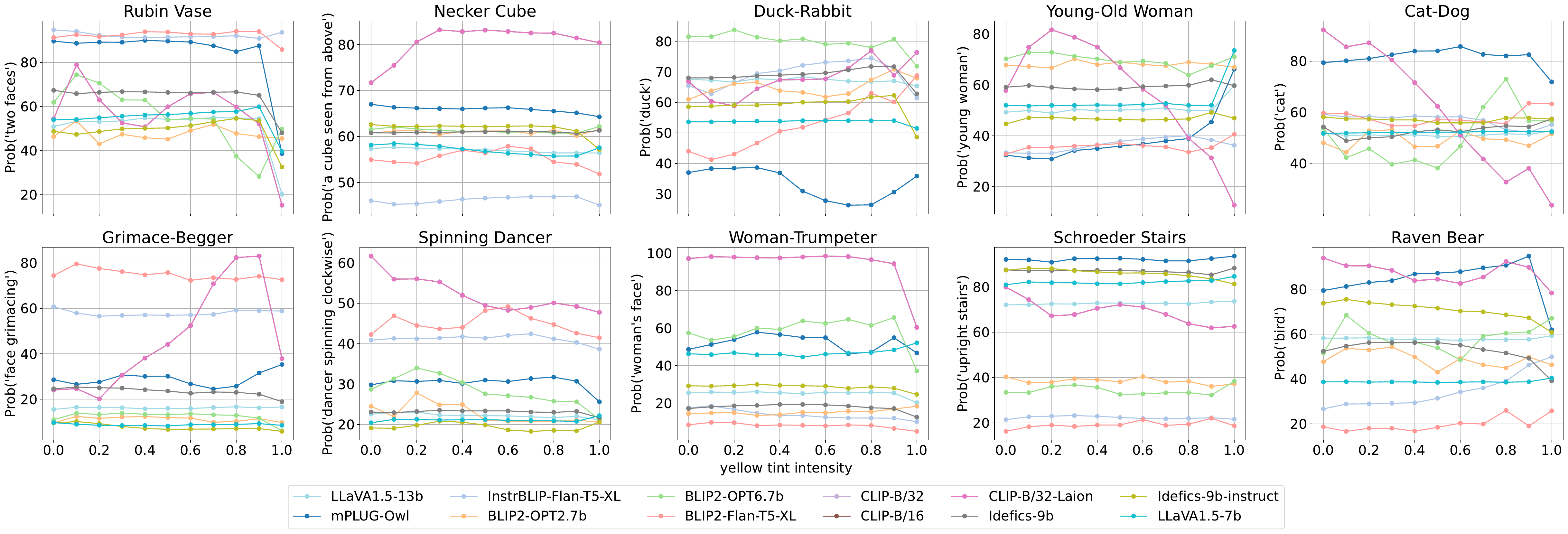}
    \caption{Yellow Tint Variation}
         \label{fig:yellow_tint}
\end{figure*}

\begin{figure*}[h]
    \centering
    \includegraphics[width=\linewidth]{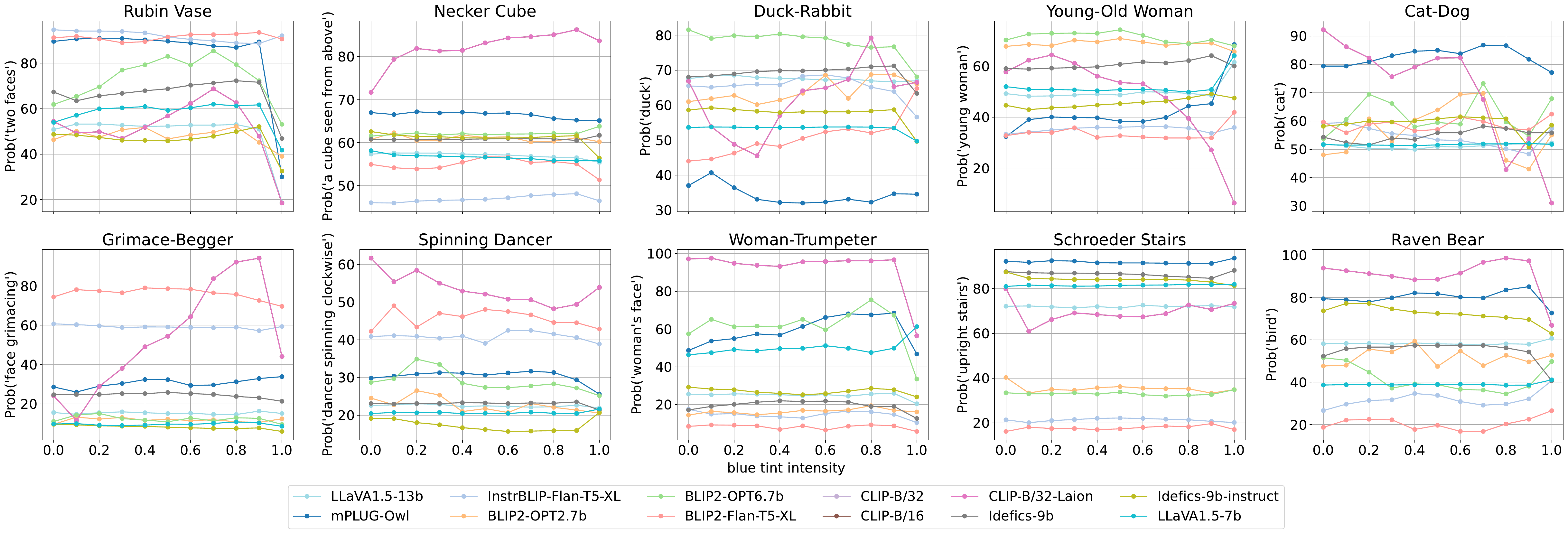}
    \caption{Blue Tint Variation}
         \label{fig:blue_tint}
\end{figure*}

\begin{figure*}[h]
    \centering
    \includegraphics[width=\linewidth]{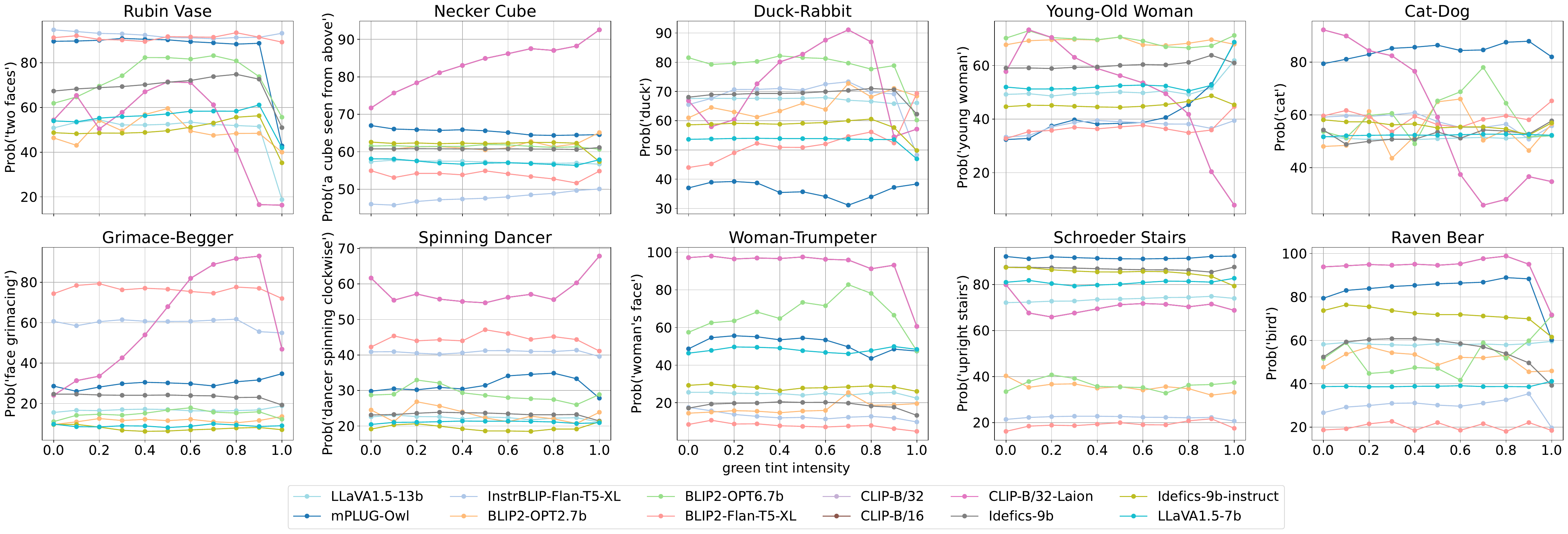}
    \caption{Green Tint Variation}
         \label{fig:green_tint}
\end{figure*}

Figures \ref{fig:red_tint}, \ref{fig:cyan_tint}, \ref{fig:magenta_tint}, \ref{fig:yellow_tint}, \ref{fig:blue_tint}, \ref{fig:green_tint} show the indiividual variations of each model for each image category based on tint variations. We find limited effect in preferences with highest variability observed by the CLIP-B/32 variants. Interestingly, most models seem to show same preferences when full tint is applied, indicating a monochrome image - hence the linguistic priors play a large role in model behavior as indicated by the synonym and prompt variation experiments. 

\section{Bistable Image Collection}
\label{app:collection}

\begin{figure*}[ht]
  \centering
  \begin{subfigure}[b]{0.11\linewidth}
    \includegraphics[width=\linewidth]{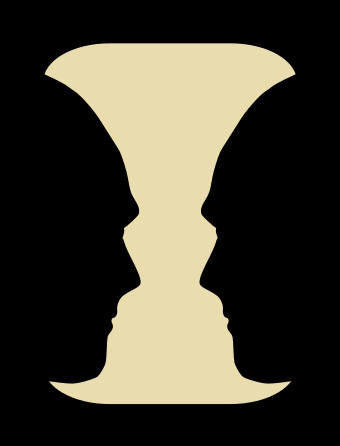}
  \end{subfigure} \hspace{12pt}
  \begin{subfigure}[b]{0.15\linewidth}
    \includegraphics[width=\linewidth]{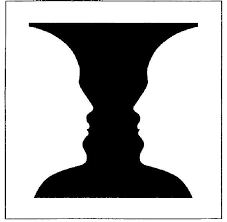}
  \end{subfigure} \hspace{12pt}
  \begin{subfigure}[b]{0.16\linewidth}
    \includegraphics[width=\linewidth]{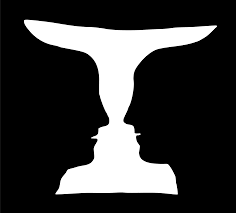}
  \end{subfigure} \hspace{12pt}
  \begin{subfigure}[b]{0.18\linewidth}
    \includegraphics[width=\linewidth]{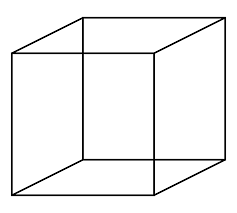}
  \end{subfigure} \hspace{12pt}
  \begin{subfigure}[b]{0.14\linewidth}
    \includegraphics[width=\linewidth]{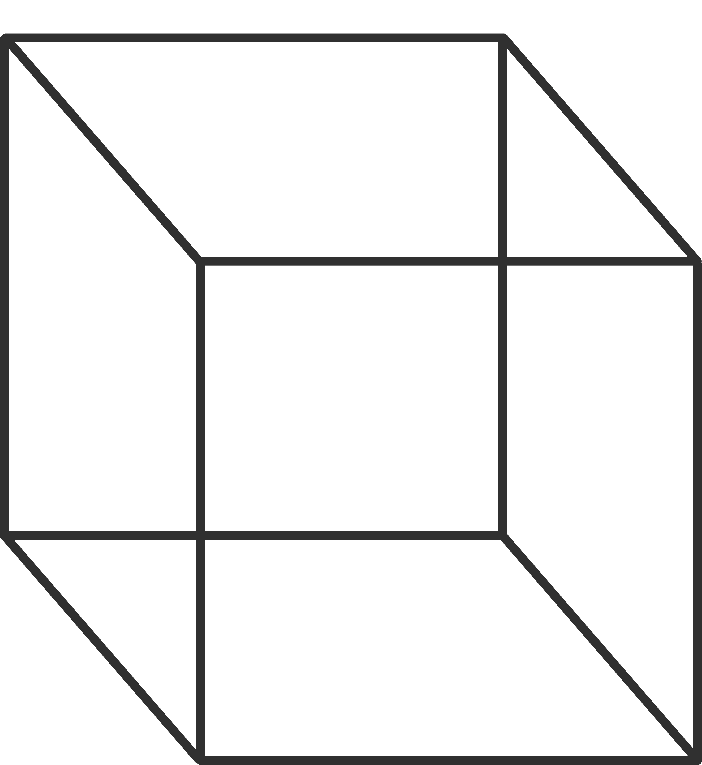}
  \end{subfigure}
  \caption{Rubin Vase illusions (interpretations: ["vase", "two faces"]) and Necker Cube illusions (interpretations: ["a cube seen from below", "a cube seen from above"]).}
  \label{fig:rubin_vase_org_images}
\end{figure*}

\begin{figure*}[ht]
  \centering
  \begin{subfigure}[b]{0.15\linewidth}
    \includegraphics[width=\linewidth]{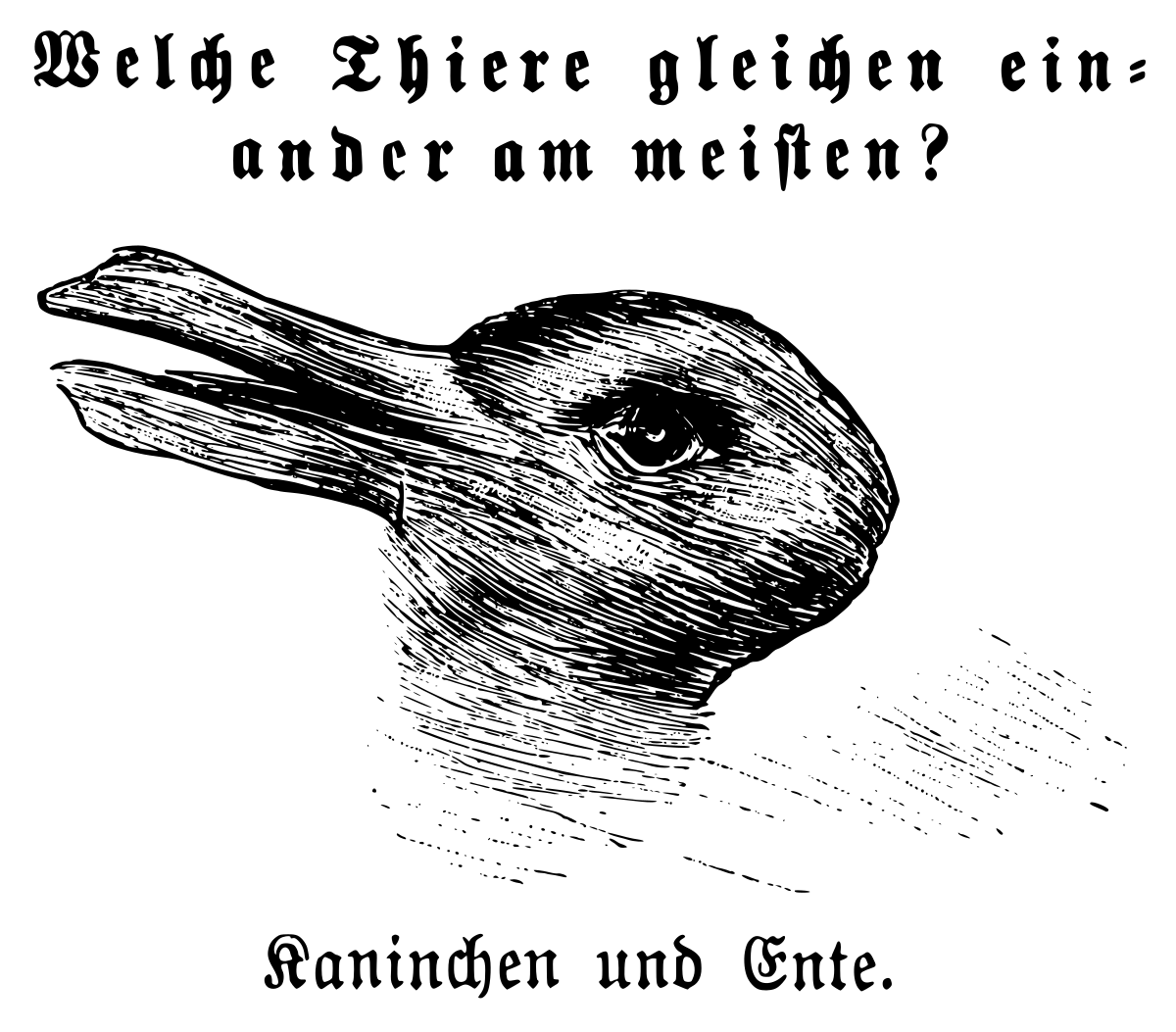}
  \end{subfigure} \hspace{12pt}
  \begin{subfigure}[b]{0.12\linewidth}
    \includegraphics[width=\linewidth]{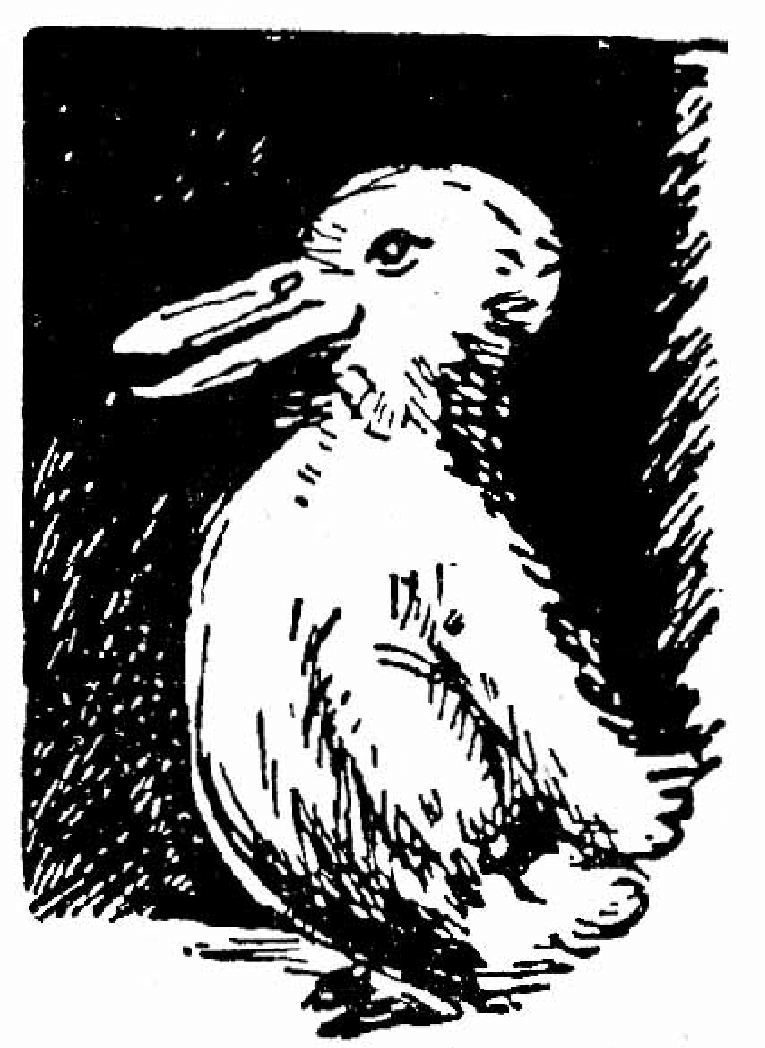}
  \end{subfigure} \hspace{12pt}
  \begin{subfigure}[b]{0.17\linewidth}
    \includegraphics[width=\linewidth]{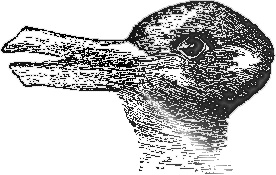}
  \end{subfigure} \hspace{12pt}
  \begin{subfigure}[b]{0.15\linewidth}
    \includegraphics[width=\linewidth]{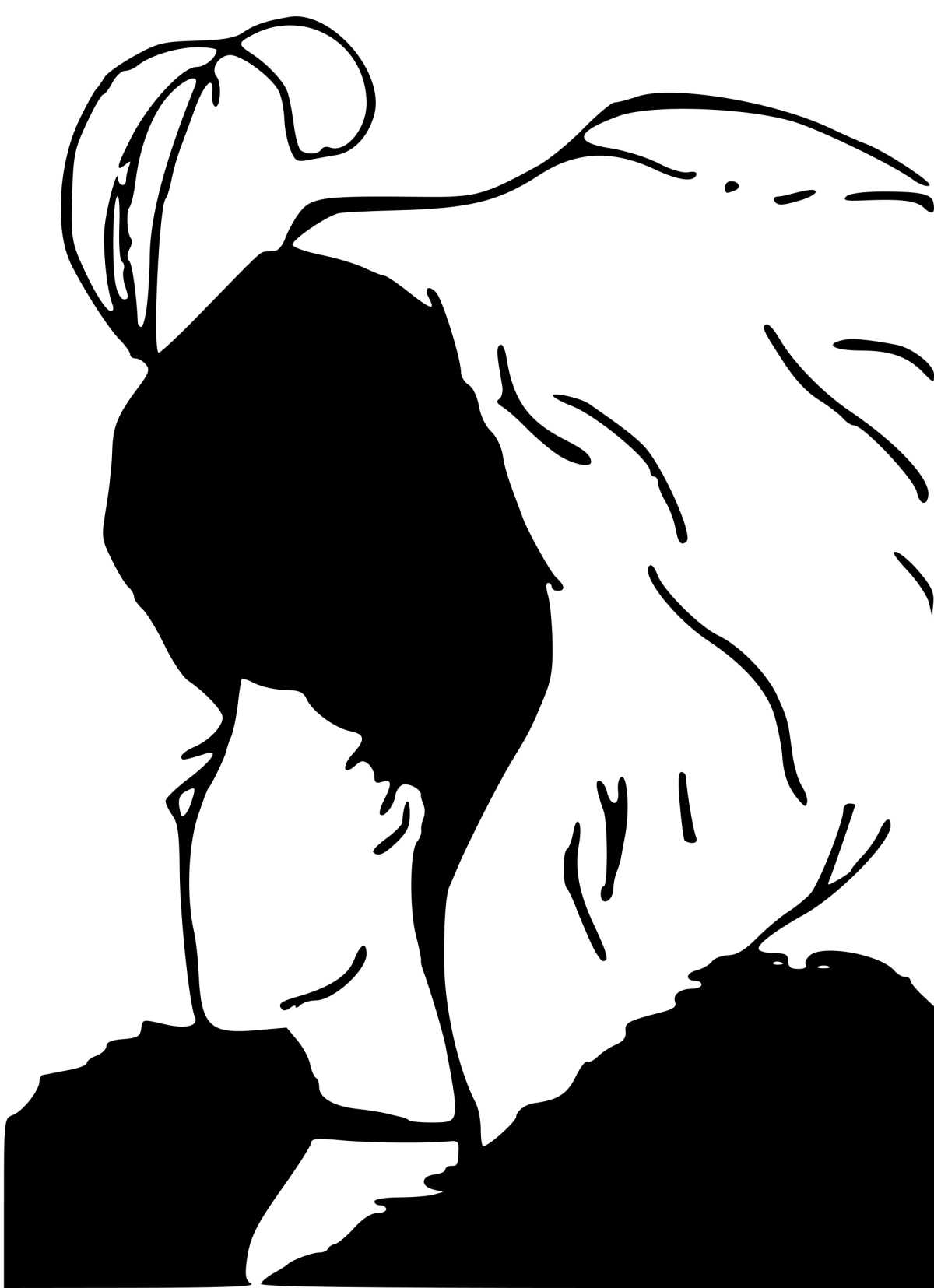}
  \end{subfigure} \hspace{12pt}
  \begin{subfigure}[b]{0.15\linewidth}
    \includegraphics[width=\linewidth]{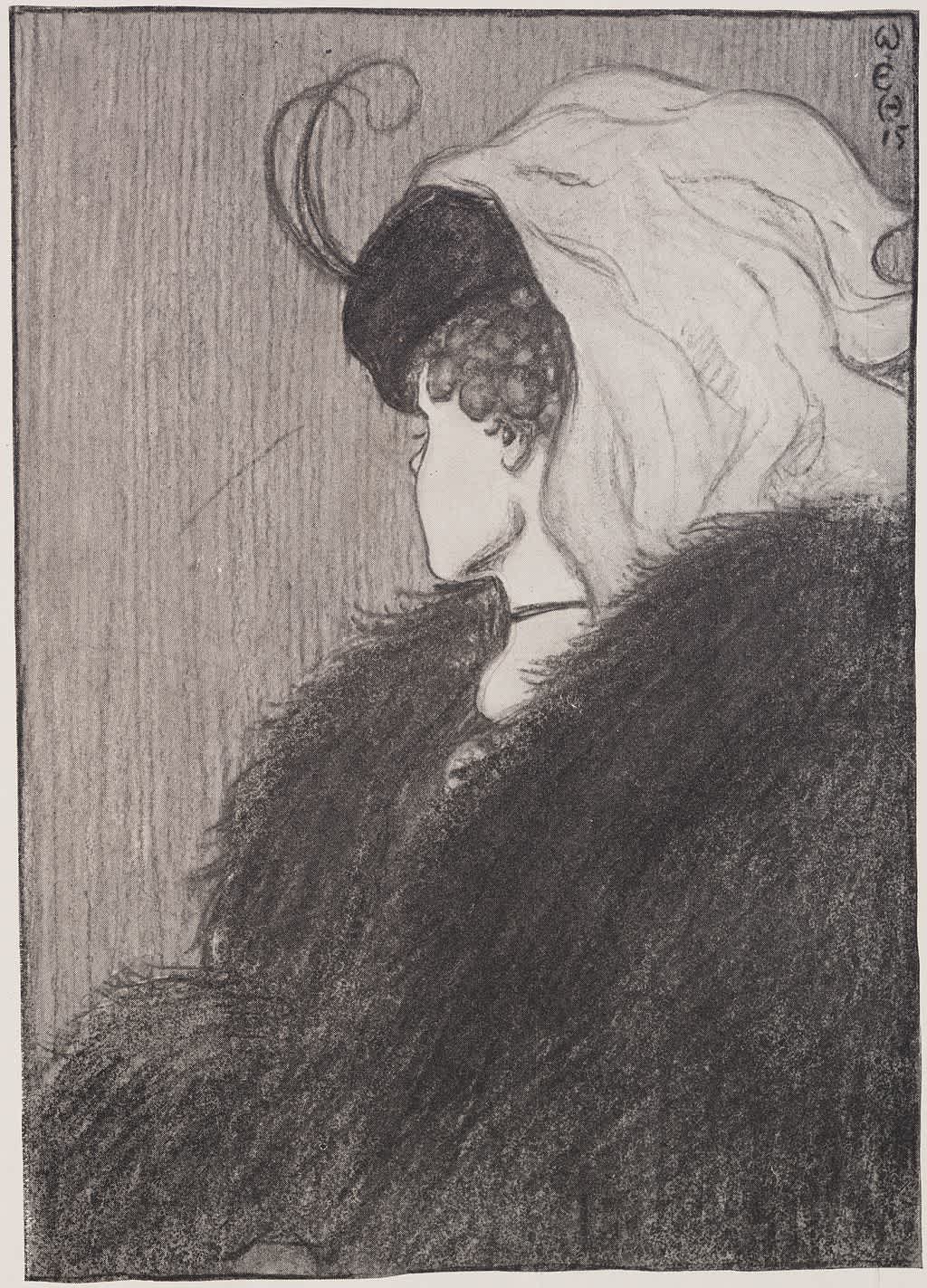}
  \end{subfigure}
  \caption{Duck-Rabbit illusion (interpretations: ["duck", "rabbit"]) and Young-Old Woman illusion (interpretations: ["young woman", "old woman"]).}
  \label{fig:duck_rabbit_and_young_old_org_images}
\end{figure*}

\begin{figure*}[ht]
  \centering
    \begin{subfigure}[b]{0.15\linewidth}
    \includegraphics[width=\linewidth]{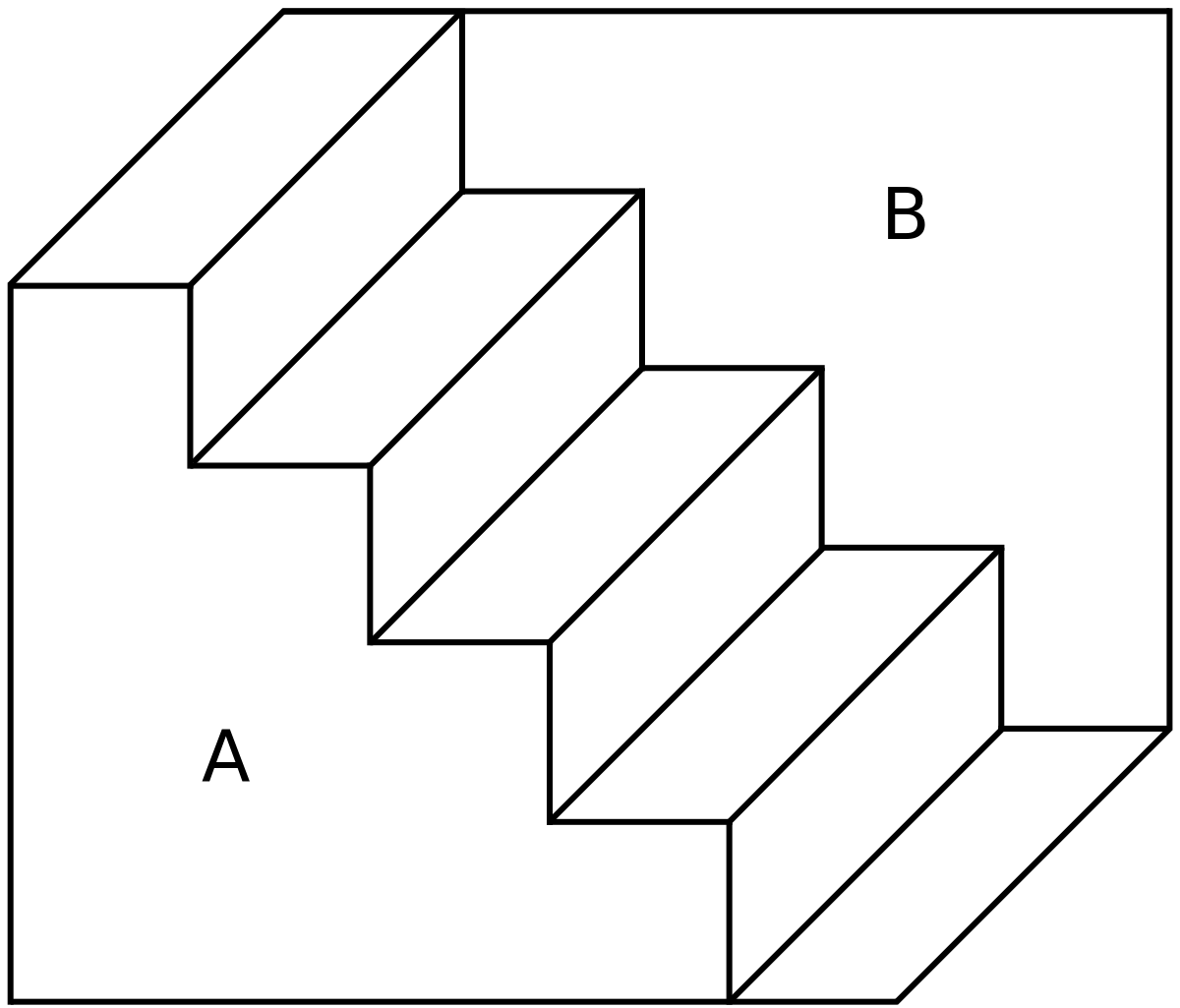}
  \end{subfigure} \hspace{12pt}
    \begin{subfigure}[b]{0.15\linewidth}
    \includegraphics[width=\linewidth]{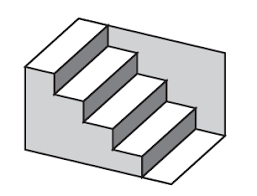}
  \end{subfigure} \hspace{12pt}
  \begin{subfigure}[b]{0.15\linewidth}
    \includegraphics[width=\linewidth]{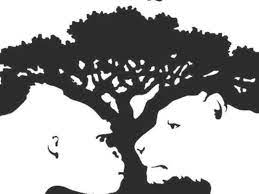}
  \end{subfigure} \hspace{12pt}
  \begin{subfigure}[b]{0.15\linewidth}
    \includegraphics[width=\linewidth]{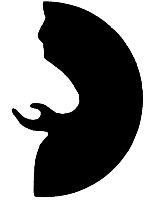}
  \end{subfigure}
  \caption{Shroeder Stairs illusion (interpretations: ["upright stairs", "sideways stairs"]), Lion-Gorilla-Tree illusion (interpretations: ["lion and gorilla", "tree"]) and Grimace-Begger illusion (interpretations: ["grimace", "beggar"]).}
  \label{fig:shroeder_lion_gorilla_tree_grimace_begger_org_images}
\end{figure*}

\begin{figure*}[ht]
  \centering
  \begin{subfigure}[b]{0.15\linewidth}
    \includegraphics[width=\linewidth]{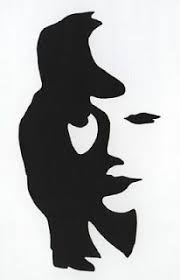}
  \end{subfigure} \hspace{12pt}
  \begin{subfigure}[b]{0.15\linewidth}
    \includegraphics[width=\linewidth]{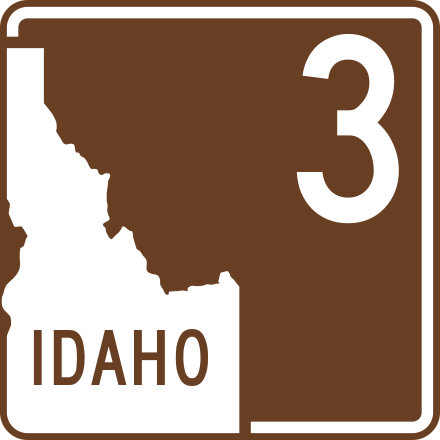}
  \end{subfigure} \hspace{12pt}
  \begin{subfigure}[b]{0.15\linewidth}
    \includegraphics[width=\linewidth]{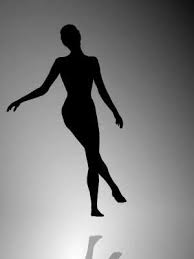}
  \end{subfigure} \hspace{12pt}
  \begin{subfigure}[b]{0.15\linewidth}
    \includegraphics[width=\linewidth]{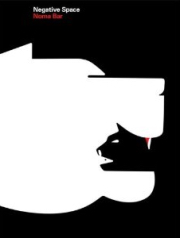}
  \end{subfigure} \hspace{12pt}
  \begin{subfigure}[b]{0.15\linewidth}
    \includegraphics[width=\linewidth]{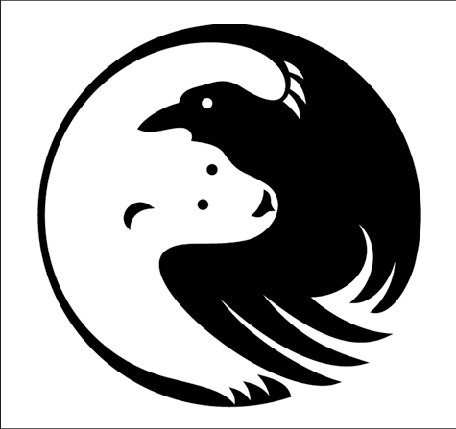}
  \end{subfigure}
  \caption{Various illusions from left to right: Woman-Trumpeter (interpretations: ["woman's face", "saxophonist"], Idaho-Face (interpretations: ["the state of Idaho", "face"]), Spinning Dancer  (interpretations: ["dancer spinning clockwise", "dancer spinning counter-clockwise"]), Cat-Dog  (interpretations: ["cat", "dog"]), and Raven-Bear  (interpretations: ["bird", "bear"]).}
  \label{fig:various_org_images}
\end{figure*}

We present the original images on our dataset, without any visual manipulations in figures \ref{fig:rubin_vase_org_images}, \ref{fig:duck_rabbit_and_young_old_org_images}, \ref{fig:shroeder_lion_gorilla_tree_grimace_begger_org_images}, \ref{fig:various_org_images}. 


\section{Generative Examples}
\label{app:qualitative}
We present a list of all the generations from the models prompted with "describe the image" in figures with the exception of few question based prompts: ``What is the orientation of the staircase/cube?" for the Shroeder stairs and Necker Cube illusions, and ``What is the dancer's spinning direction?" for `Spinning Dancer'. 
\begin{figure*}
    \centering
    \includegraphics[width=\linewidth]{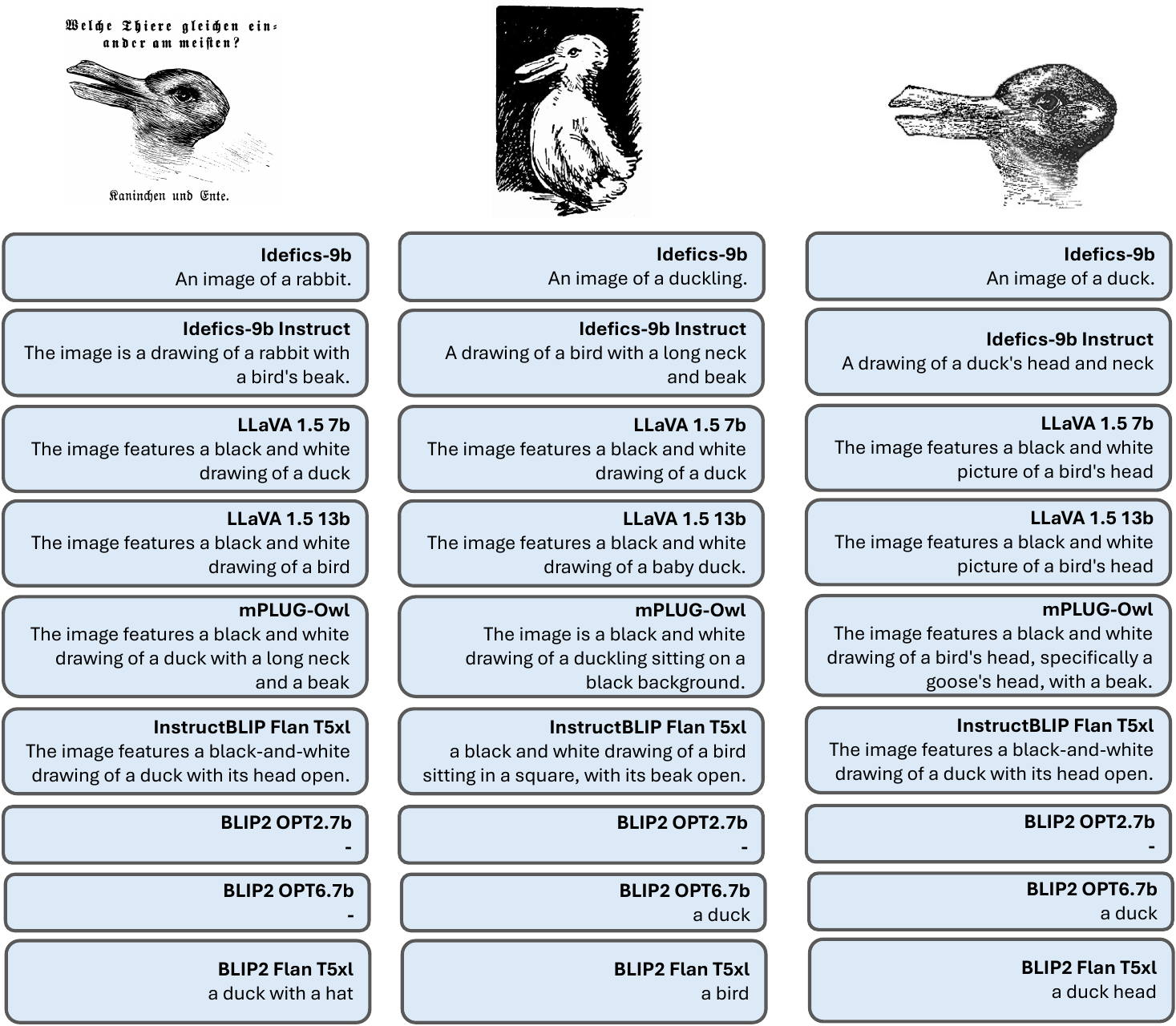}
    \caption{Duck-Rabbit generative examples}
    \label{fig:duck_rabbit_gen}
\end{figure*}

\begin{figure*}
    \centering
    \includegraphics[width=\linewidth]{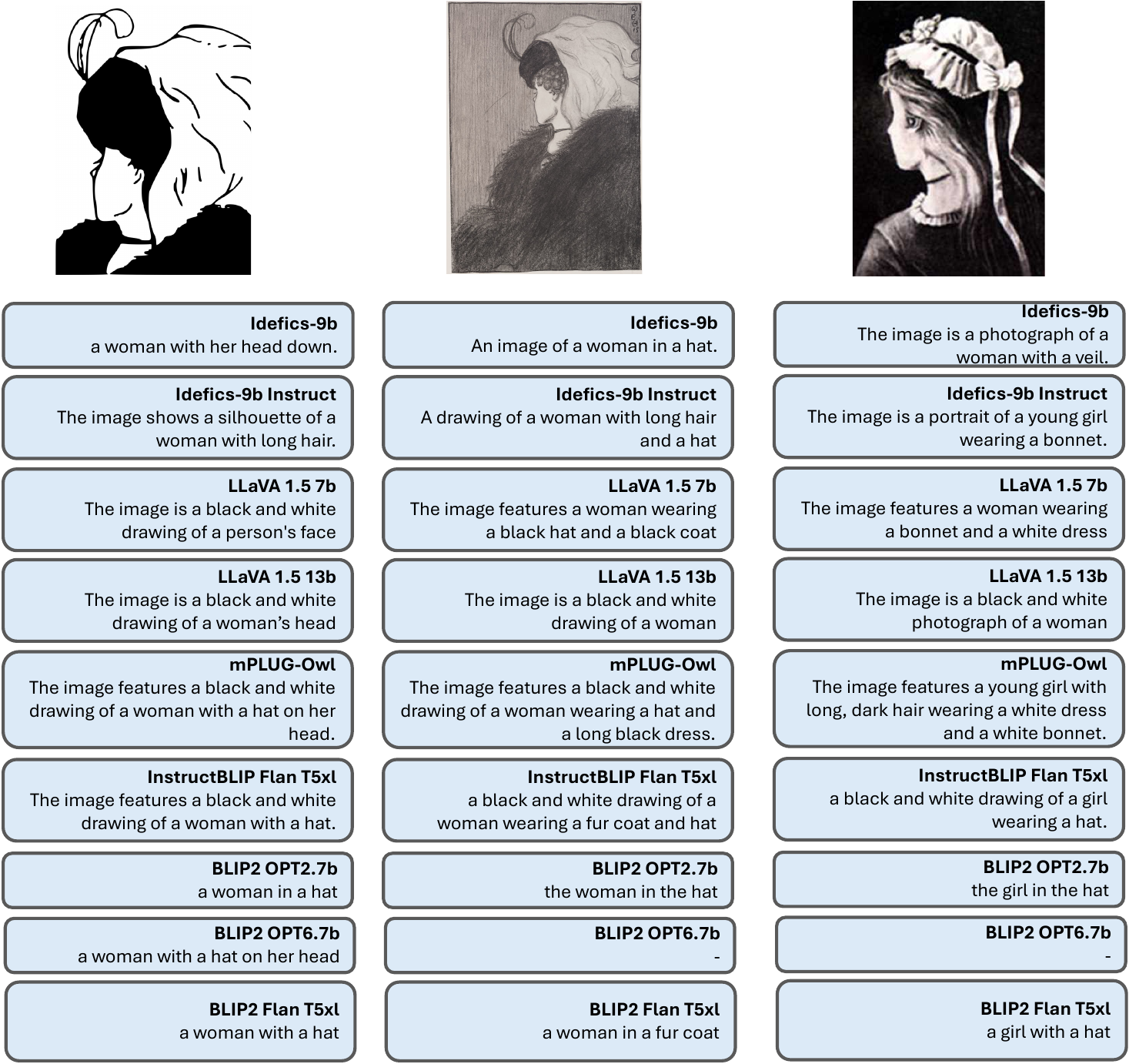}
    \caption{Young Old woman generative examples}
    \label{fig:young_old_gen}
\end{figure*}

\begin{figure*}
    \centering
    \includegraphics[width=\linewidth]{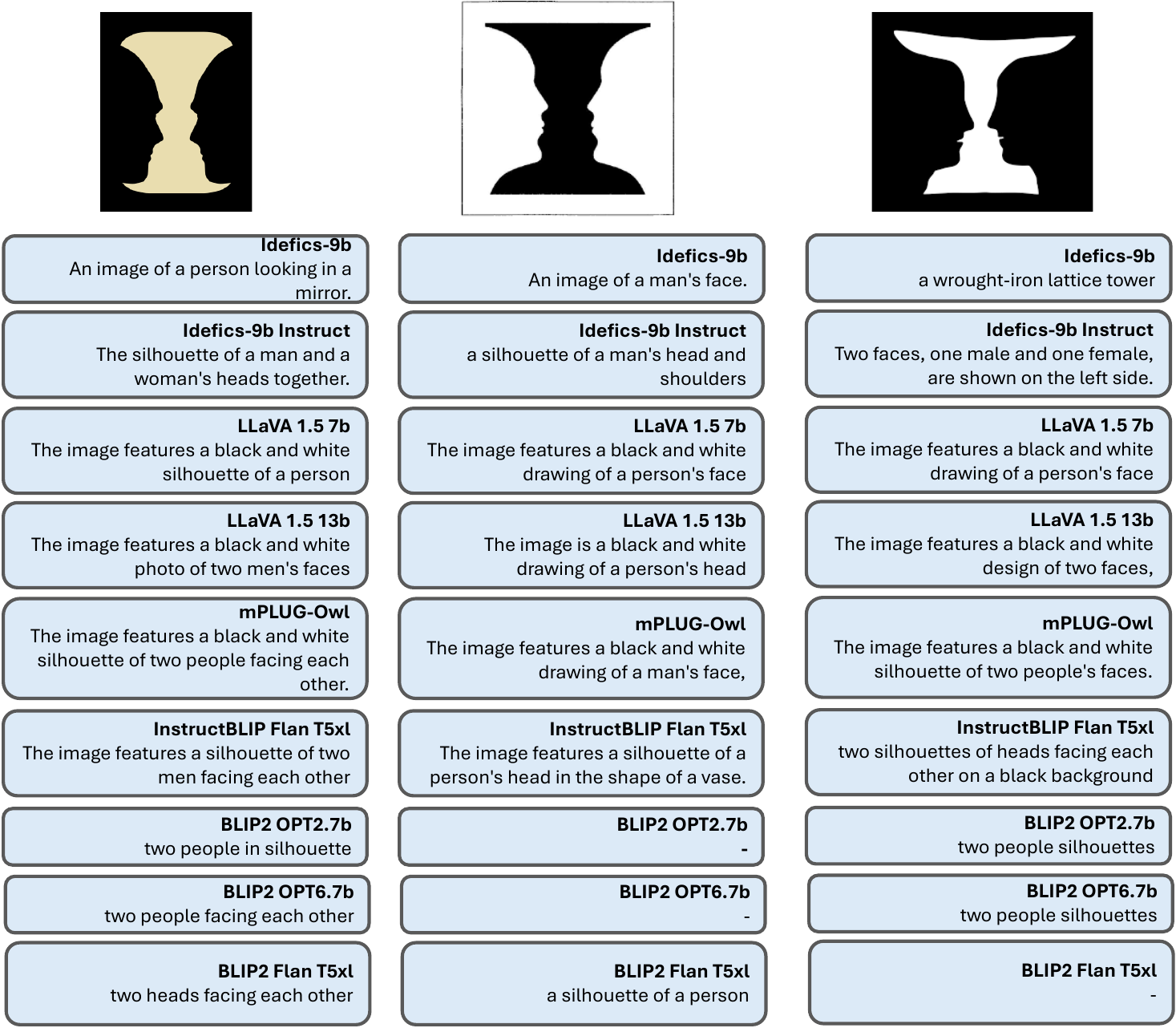}
    \caption{Vase-Faces woman generative examples}
    \label{fig:vase_faces_gen}
\end{figure*}

\begin{figure*}
    \centering
    \includegraphics[width=\linewidth]{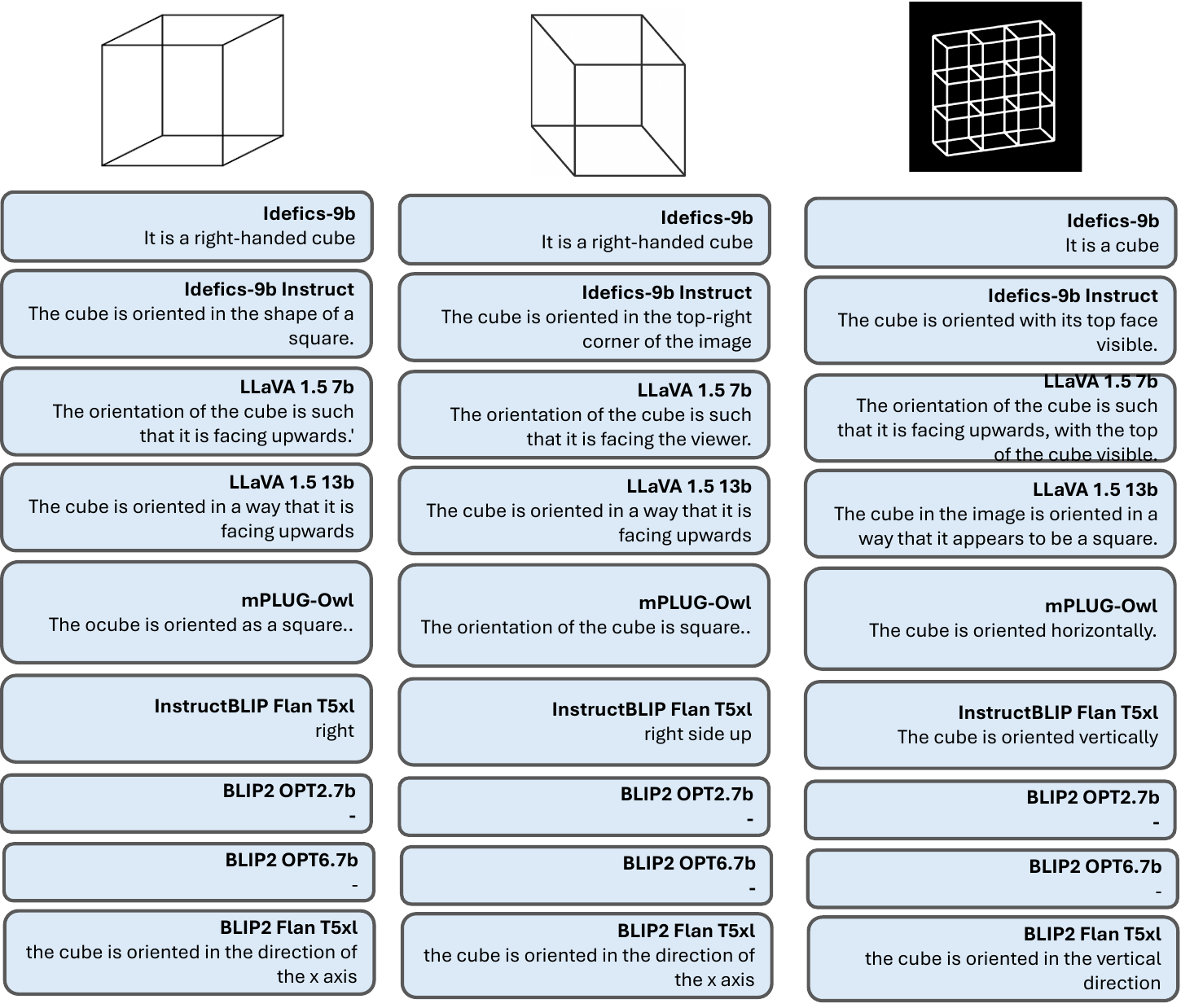}
    \caption{Necker-Cube generative examples on question ``What is the orientation of the cube?"}
    \label{fig:necker_cube_gen}
\end{figure*}

\begin{figure*}
    \centering
    \includegraphics[width=\linewidth]{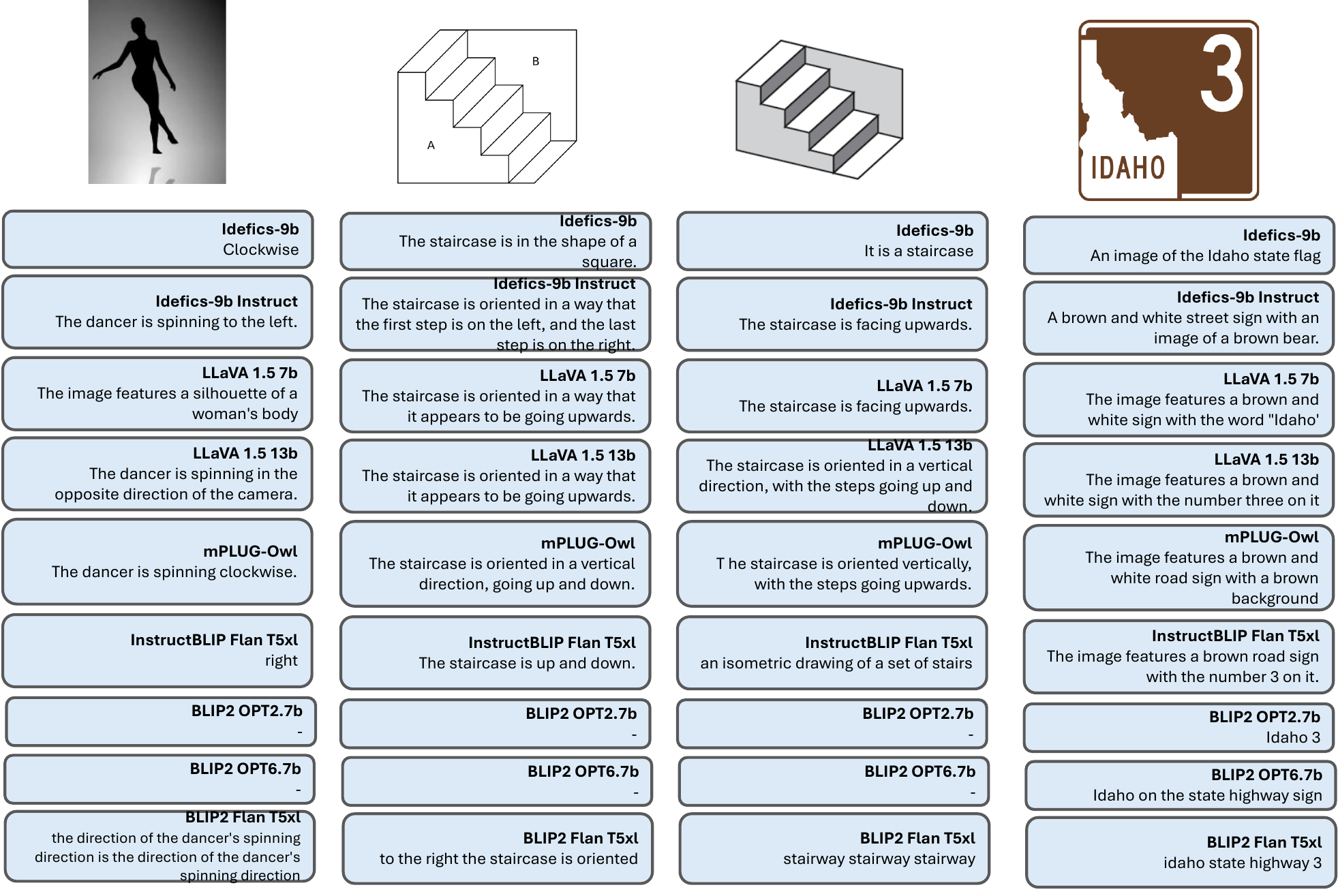}
    \caption{Spinning Dancer results on question ``What is the dancer's spinning direction?", Shroeder Stairs on ``What is the orientation of the stairs" and Idaho-Face on "describe the image".}
    \label{fig:dancer_stairs_idaho_gen}
\end{figure*}

\end{document}